\documentclass[preprint,authoryear,12pt]{elsarticle}

\usepackage{amssymb}
\usepackage{amsmath}
\usepackage{graphicx}
\usepackage{subcaption}
\usepackage{xspace}
\usepackage{bm}
\usepackage[colorlinks=true,citecolor=blue,urlcolor=black]{hyperref}

\usepackage[colorinlistoftodos]{todonotes}
\usepackage{lineno}

\usepackage{booktabs}
\usepackage{array}
\usepackage[flushleft]{threeparttable}

\newcommand{\eqfs}{\textrm{ .}}

\newcommand{\eqcm}{\textrm{ ,}}

\newcommand{\quot}[1]{``#1''}

\newcommand{\ignore}[1]{} 

\begin{document}

\begin{frontmatter}

\title{Efficient Multi-Scale 3D CNN with fully connected CRF for Accurate Brain Lesion Segmentation}

\author[ICL]{Konstantinos Kamnitsas}
\author[ICL]{Christian Ledig}
\author[UDA,WB]{Virginia F.J. Newcombe}
\author[UDA]{Joanna P. Simpson}
\author[UDA]{Andrew D. Kane}
\author[UDA,WB]{David K. Menon}
\author[ICL]{\\Daniel Rueckert}
\author[ICL]{Ben Glocker}

\address[ICL]{Biomedical Image Analysis Group, Imperial College London, UK}
\address[UDA]{University Division of Anaesthesia, Department of Medicine, Cambridge University, UK}
\address[WB]{Wolfson Brain Imaging Centre, Cambridge University, UK}

\begin{abstract}
We propose a dual pathway, 11-layers deep, three-dimensional Convolutional Neural Network for the challenging task of brain lesion segmentation. The devised architecture is the result of an in-depth analysis of the limitations of current networks proposed for similar applications. To overcome the computational burden of processing 3D medical scans, we have devised an efficient and effective dense training scheme which joins the processing of adjacent image patches into one pass through the network while automatically adapting to the inherent class imbalance present in the data. Further, we analyze the development of deeper, thus more discriminative 3D CNNs. In order to incorporate both local and larger contextual information, we employ a dual pathway architecture that processes the input images at multiple scales simultaneously. For post-processing of the network's soft segmentation, we use a 3D fully connected Conditional Random Field which effectively removes false positives. Our pipeline is extensively evaluated on three challenging tasks of lesion segmentation in multi-channel MRI patient data with traumatic brain injuries, brain tumors, and ischemic stroke. We improve on the state-of-the-art for all three applications, with top ranking performance on the public benchmarks BRATS 2015 and ISLES 2015. Our method is computationally efficient, which allows its adoption in a variety of research and clinical settings. The source code of our implementation is made publicly available.
\end{abstract}


\end{frontmatter}



\section{Introduction}

Segmentation and the subsequent quantitative assessment of lesions in medical images provide valuable information for the analysis of neuropathologies and are important for planning of treatment strategies, monitoring of disease progression and prediction of patient outcome. For a better understanding of the pathophysiology of diseases, quantitative imaging can reveal clues about the disease characteristics and effects on particular anatomical structures. For example, the associations of different lesion types, their spatial distribution and extent with acute and chronic sequelae after traumatic brain injury (TBI) are still poorly understood (\cite{Maas2015}). However, there is growing evidence that quantification of lesion burden may add insight into the functional outcome of patients (\cite{Ding2008,Moen2012}). Additionally, exact locations of injuries relate to particular deficits depending on the brain structure that is affected (\cite{lehtonen2005neuropsychological, Warner2010d, Sharp2011}). This is in line with estimates that functional deficits caused by stroke are associated with the extent of damage to particular parts of the brain (\cite{carey2013beyond}). Lesion burden is commonly quantified by means of volume and number of lesions, biomarkers that have been shown to be related to cognitive deficits. For example, volume of white matter lesions (WML) correlates with cognitive decline and increased risk of dementia (\cite{Ikram2010}). In clinical research on multiple sclerosis (MS), lesion count and volume are used to analyse disease progression and effectiveness of pharmaceutical treatment (\cite{Rovira2008,Kappos2007}). Finally, accurate delineation of the pathology is important in the case of brain tumors, where estimation of the relative volume of a tumor's sub-components is required for planning radiotherapy and treatment follow-up (\cite{Wen2010}).

The quantitative analysis of lesions requires accurate lesion segmentation in multi-modal, three-dimensional images which is a challenging task for a number of reasons. The heterogeneous appearance of lesions including the large variability in location, size, shape and frequency make it difficult to devise effective segmentation rules.
It is thus highly non-trivial to delineate contusions, edema and haemorrhages in TBI (\cite{Irimia2012a}), or sub-components of brain tumors such as proliferating cells and necrotic core (\cite{Menze2014}). The arguably most accurate segmentation results can be obtained through manual delineation by a human expert which is tedious, expensive, time-consuming, impractical in larger studies, and introduces inter-observer variability. Additionally, for deciding whether a particular region is part of a lesion multiple image sequences with varying contrasts need to be considered, and the level of expert knowledge and experience are important factors that impact segmentation accuracy. Hence, in clinical routine often only qualitative, visual inspection, or at best crude measures like approximate lesion volume and number of lesions are used (\cite{Yuh2012,Wen2010}). In order to capture and better understand the complexity of brain pathologies it is important to conduct large studies with many subjects to gain the statistical power for drawing conclusions across a whole patient population. The development of accurate, automatic segmentation algorithms has therefore become a major research focus in medical image computing with the potential to offer objective, reproducible, and scalable approaches to quantitative assessment of brain lesions.

\begin{figure}[h] 
\centering
\begin{subfigure}[b]{0.225\textwidth}
	\centering
	\includegraphics[clip=true, trim=10pt 0pt 10pt 0pt, width=1.\textwidth]{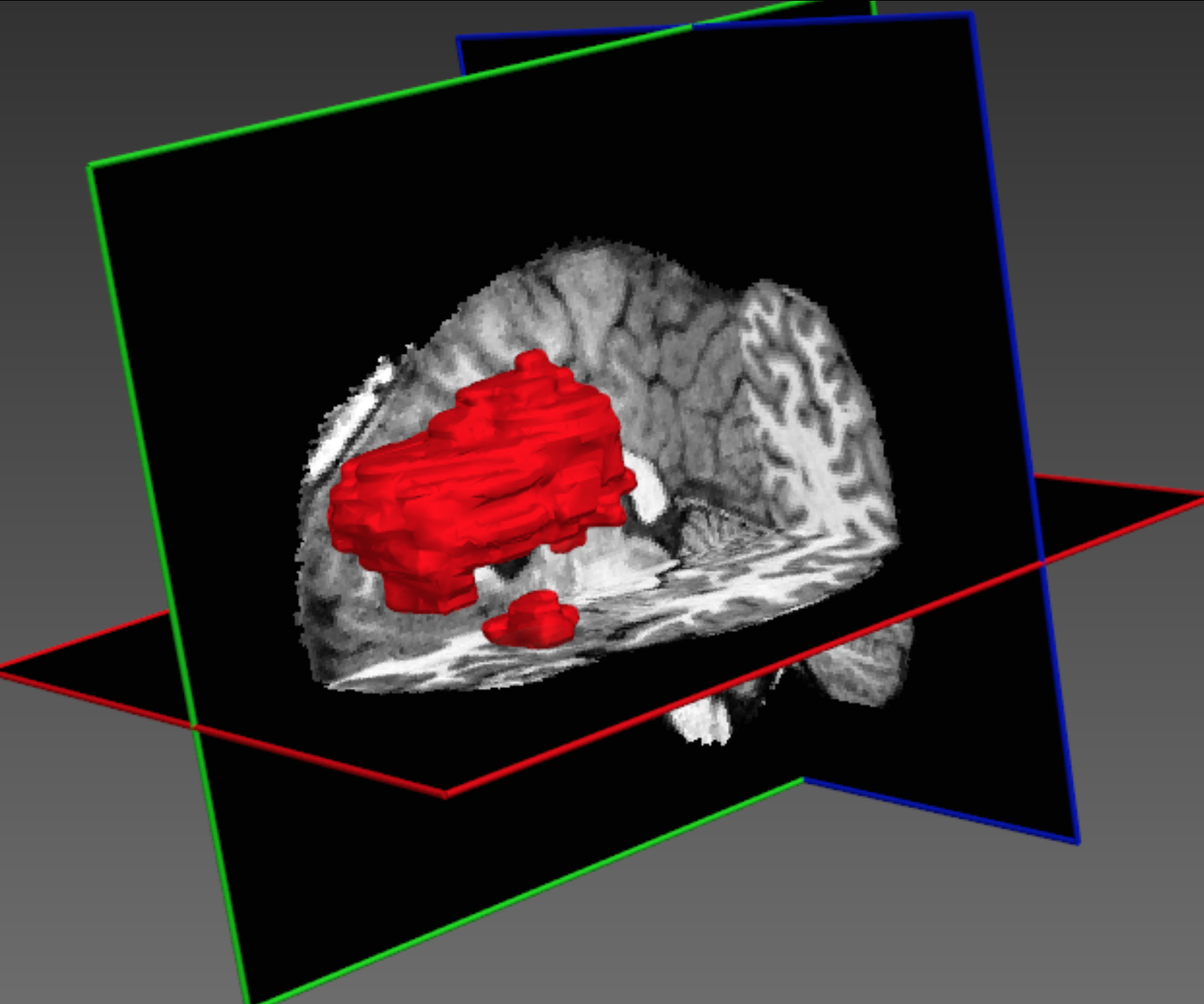}
	\caption{}
	\label{fig:3dLesionBig}
\end{subfigure}
\begin{subfigure}[b]{0.225\textwidth}
	\centering
	\includegraphics[clip=true, trim=10pt 0pt 10pt 0pt, width=1.\textwidth]{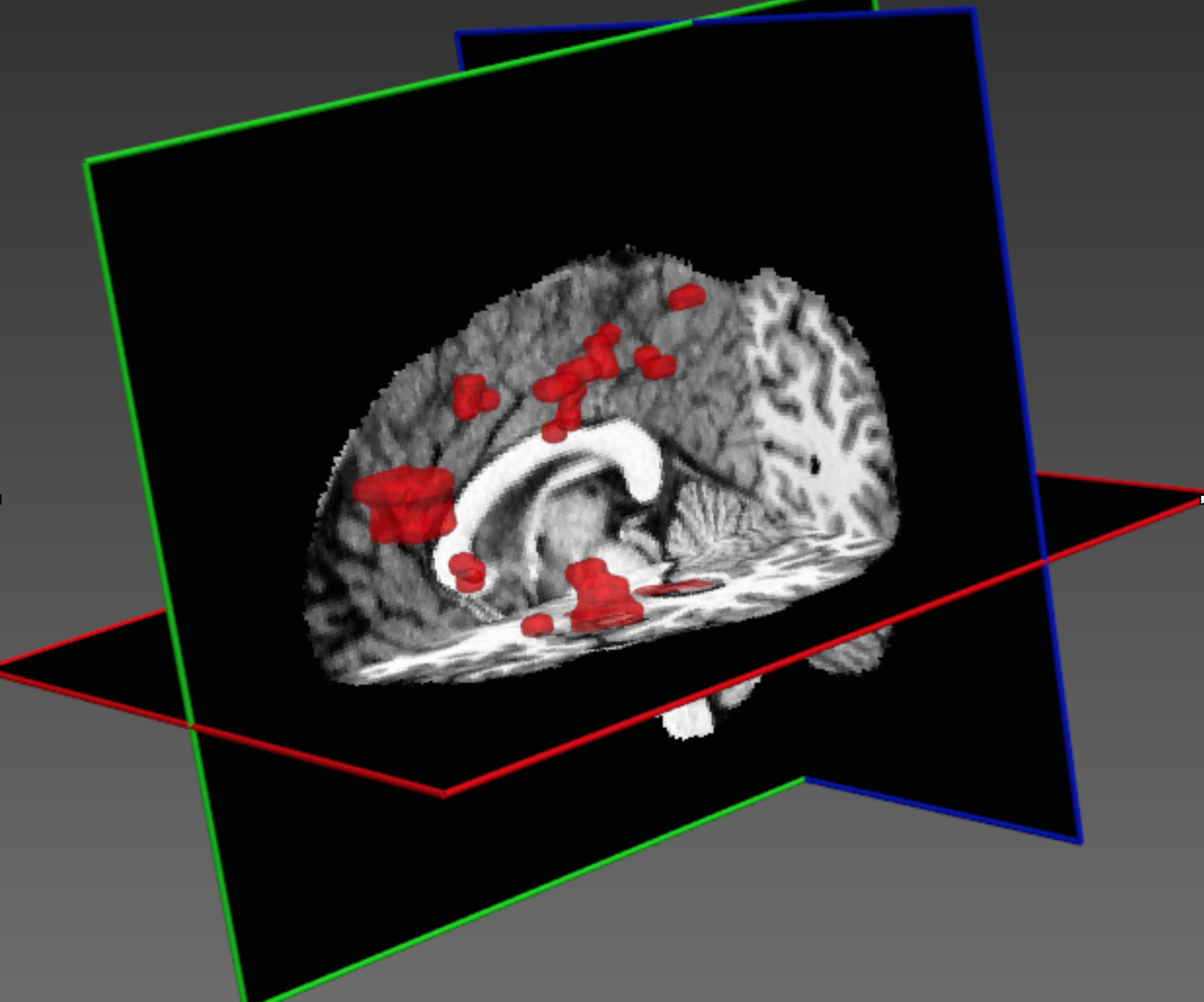}
	\caption{}
	\label{fig:3dLesionSmalls}
\end{subfigure}
\begin{subfigure}[b]{0.225\textwidth}
	\centering
	\includegraphics[clip=true, trim=10pt 20pt 10pt 20pt, width=0.7\textwidth]{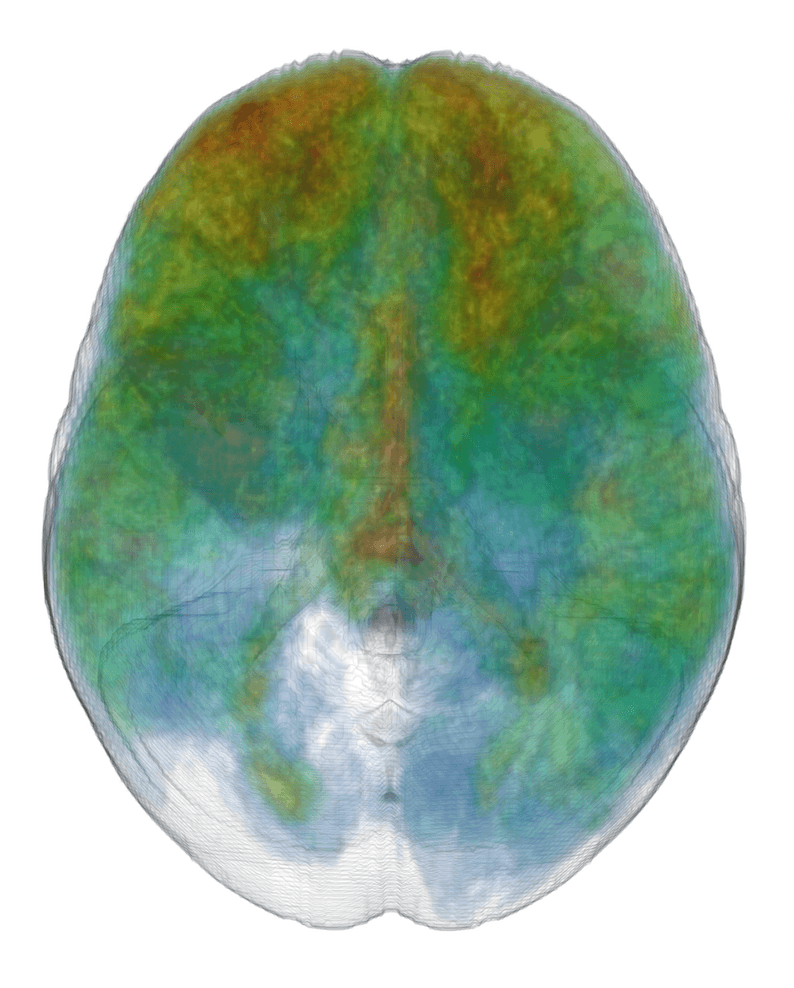}
	\caption{}
	\label{fig:spatialMap}
\end{subfigure}
\begin{subfigure}[b]{0.22\textwidth}
	\centering
	\includegraphics[clip=true, trim=0pt 190pt 0pt 0pt, width=1.0\textwidth]{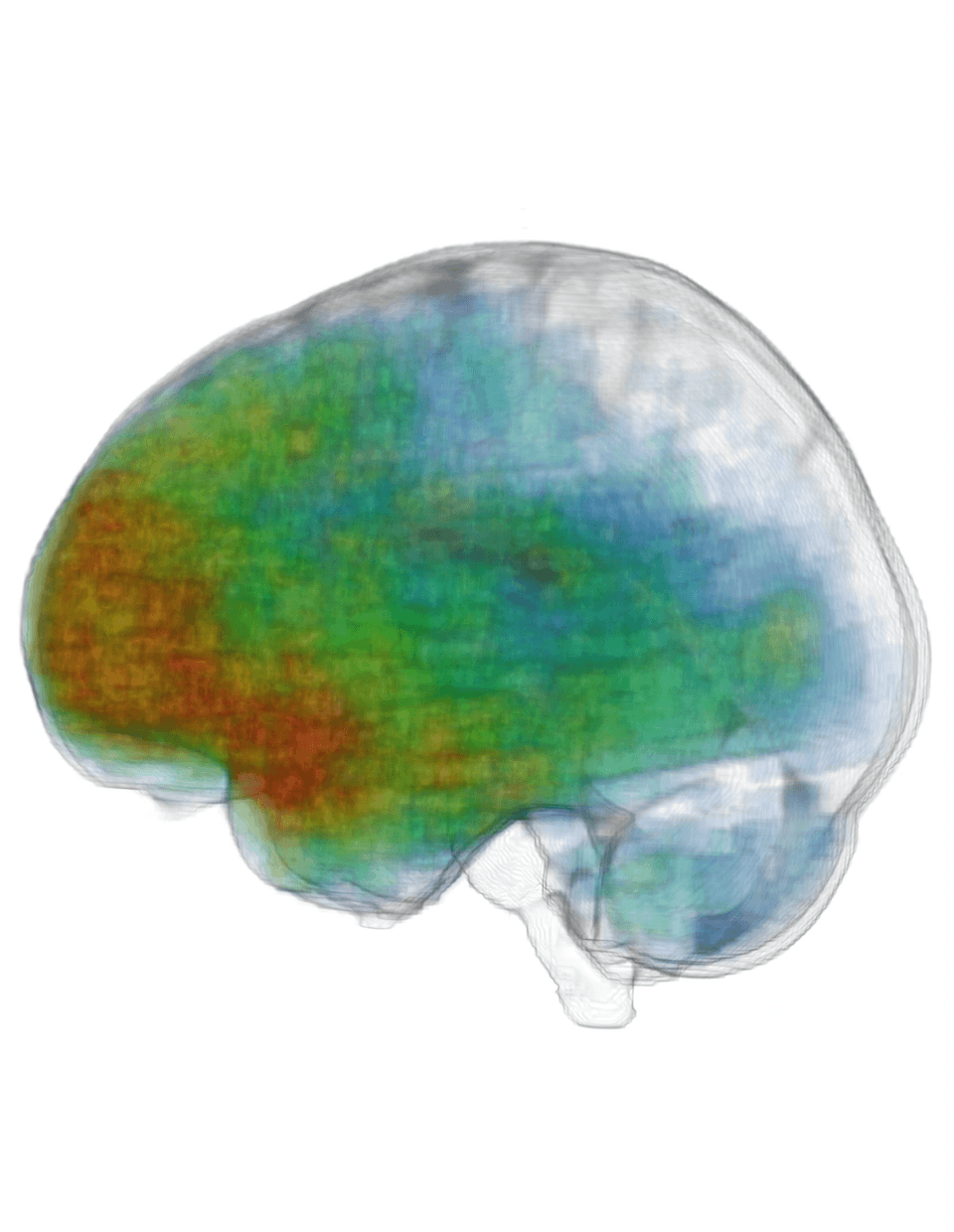}
	\caption{}
	\label{fig:spatialMapSide}
\end{subfigure}
\\[1ex] 
\centering
\begin{subfigure}[b]{1.0\textwidth}
\centering
	\includegraphics[clip=true, trim=0pt 0pt 0pt 0pt, width=1.0\textwidth]{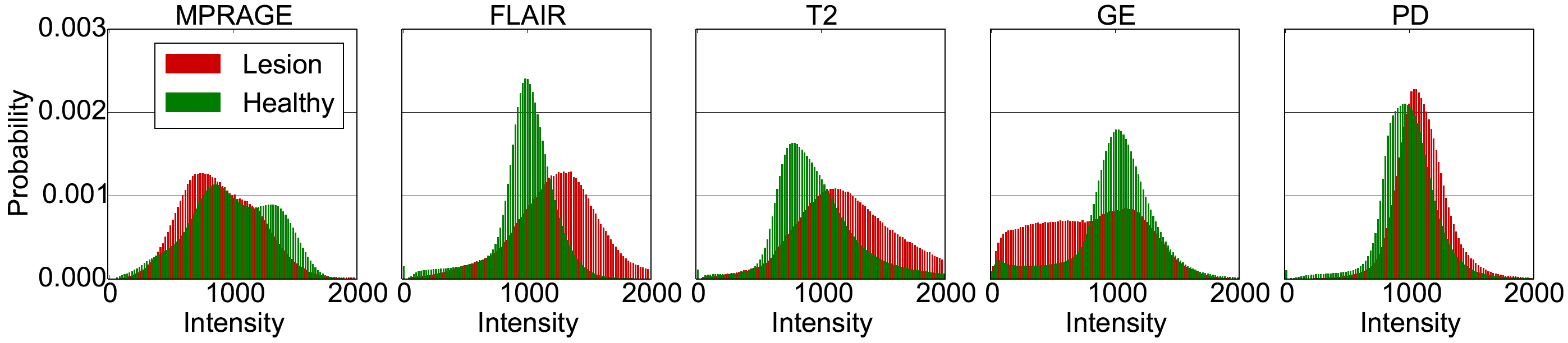}
	\caption{}
	\label{fig:histInt}
\end{subfigure}

\caption{Heterogeneous appearance of TBI lesions poses challenges in devising discriminative models. Lesion size varies significantly with both large, focal and small, diffused lesions (a,b). Alignment of manual lesion segmentations reveals the wide spatial distribution of lesions in (c,d) with some areas being more likely than others. (e) shows the average of the normalized intensity histograms of different MR channels over all the TBI cases in our database, for healthy (green) and injured (red) tissue. One can observe a large overlap between the distributions of healthy and non-healthy tissue.}
\label{fig:tbiChallenges}
\end{figure}

Figure~\ref{fig:tbiChallenges} illustrates some of the challenges that arise when devising a computational approach for the task of automatic lesion segmentation. The figure summarizes statistics and shows examples of brain lesions in the case of TBI, but is representative of other pathologies such as brain tumors and ischemic stroke. Lesions can occur at multiple sites, with varying shapes and sizes, and their image intensity profiles largely overlap with non-affected, healthy parts of the brain or lesions which are not in the focus of interest. For example, stroke and MS lesions have a similar hyper-intense appearance in FLAIR sequences as other WMLs (\cite{Mitra2014, Schmidt2012}). It is generally difficult to derive statistical prior information about lesion shape and appearance. On the other hand, in some applications there is an expectation on the spatial configuration of segmentation labels, for example there is a hierarchical layout of sub-components in brain tumors. Ideally, a computational approach is able to adjust itself to application specific characteristics by learning from a set of a few example images.

\subsection{Related Work}

A multitude of automatic lesion segmentation methods have been proposed over the last decade, and several main categories of approaches can be identified. One group of methods poses the lesion segmentation task as an abnormality detection problem, for example by employing image registration. The early work of \cite{Prastawa2004} and more recent ones by \cite{Schmidt2012} and \cite{doyle2013Brats} align the pathological scan to a healthy atlas and lesions are detected based on deviations in tissue appearance between the patient and the atlas image. Lesions, however, may cause large structural deformations that may lead to incorrect segmentation due to incorrect registration. \cite{Gooya2011,Parisot2012} alleviate this problem by jointly solving the segmentation and registration tasks. \cite{Liu2014} showed that registration together with a low-rank decomposition gives as a by-product the abnormal structures in the sparse components, although, this may not be precise enough for detection of small lesions. Abnormality detection has also been proposed within image synthesis works. Representative approaches are those of \cite{Weiss2013} using dictionary learning and \cite{Ye2013a} using a patch-based approach. The idea is to synthesize pseudo-healthy images that when compared to the patient scan allow to highlight abnormal regions. In this context, \cite{cardoso15} present a generative model for image synthesis that yields a probabilistic segmentation of abnormalities. Another unsupervised technique is proposed by \cite{Erihov2015}, a saliency-based method that exploits brain asymmetry in pathological cases. A common advantage of the above methods is that they do not require a training dataset with corresponding manual annotations. In general, these approaches are more suitable for detecting lesions rather than accurately segmenting them.

Some of the most successful, supervised segmentation methods for brain lesions are based on voxel-wise classifiers, such as Random Forests. Representative work is that of \cite{Geremia2010} on MS lesions, employing intensity features to capture the appearance of the region around each voxel. \cite{Zikic2012} combine this with a generative Gaussian Mixture Model (GMM) to obtain tissue-specific probabilistic priors (\cite{Leemput1999}). This framework was adopted in multiple works, with representative pipelines for brain tumors by \cite{tustison2013Brats} and TBI by \cite{Rao2014b}. Both works incorporate morphological and contextual features to better capture the heterogeneity of lesions. \cite{Rao2014b} also incorporate brain structure segmentation results obtained from a multi-atlas label propagation approach (\cite{Ledig2015}) to provide strong tissue-class priors to the Random Forests. \cite{tustison2013Brats} additionally use a Markov Random Field (MRF) to incorporate spatial regularization. MRFs are commonly used to encourage spatial continuity of the segmentation (\cite{Schmidt2012, Mitra2014}). Although those methods have been very successful, it appears that their modeling capabilities still have significant limitations. This is confirmed by the results of the most recent challenges \footnote{links: \href{http://braintumorsegmentation.org/}{http://braintumorsegmentation.org/}, \href{www.isles-challenge.org}{www.isles-challenge.org}}, and also by our own experience and experimentation with such approaches.  

At the same time, deep learning techniques have emerged as a powerful alternative for supervised learning with great model capacity and the ability to learn highly discriminative features for the task at hand. These features often outperform hand-crafted and pre-defined feature sets. In particular, Convolutional Neural Networks (CNNs) (\cite{LeCun1998, Krizhevsky2012}) have been applied with promising results on a variety of biomedical imaging problems. \cite{Ciresan2012} presented the first GPU implementation of a two-dimensional CNN for the segmentation of neural membranes. From the CNN based work that followed, related to our approach are the methods of \cite{zikic2014CnnBrats, Havei2015Journal, pereira2015Brats}, with the latter being the best performing automatic approach in the BRATS 2015 challenge (\cite{Menze2014}). These methods are based on 2D CNNs, which have been used extensively in computer vision applications on natural images. Here, the segmentation of a 3D brain scan is achieved by processing each 2D slice independently, which is arguably a non-optimal use of the volumetric medical image data. Despite the simplicity in the architecture, the promising results obtained by these methods indicate the potential of CNNs.

Fully 3D CNNs come with an increased number of parameters and significant memory and computational requirements. Previous work discusses problems and apparent limitations when employing a 3D CNN on medical imaging data (\cite{prasoon2013Knee, Li2014a, Roth2014}). To incorporate 3D contextual information, multiple works used 2D CNNs on three orthogonal 2D patches (\cite{prasoon2013Knee, Roth2014, lyksborg2015ensemble}). In their work for structural brain segmentation, \cite{Brebisson2015a} extracted large 2D patches from multiple scales of the image and combined them with small single-scale 3D patches, in order to avoid the memory requirements of fully 3D networks.

One of the reasons that discouraged the use of 3D CNNs is the slow inference due to the computationally expensive 3D convolutions. In contrast to the 2D/3D hybrid variants (\cite{Roth2014, Brebisson2015a}), 3D CNNs can fully exploit \textit{dense-inference} (\cite{LeCun1998,Sermanet2013}), a technique that greatly decreases inference times and which we will further discuss in section \ref{subsec:theBaseline}. By employing dense-inference with 3D CNNs, \cite{Brosch2015} and \cite{urban2014CnnBrats} reported computation times of a few seconds and approximately a minute respectively for the processing of a single brain scan. Even though the size of their developed networks was limited, a factor that is directly related to a network's representational power, their results on MS and brain tumor segmentation respectively were very promising.

Performance of CNNs is significantly influenced by the strategy for extracting training samples. A commonly adopted approach is training on image patches that are equally sampled from each class. This, however, biases the classifier towards rare classes and may result in over-segmentation. To counter this, \cite{Ciresan2013} proposes to train a second CNN on samples with a class distribution close to the real one, but oversample pixels that were incorrectly classified in the first stage. A secondary training stage was also suggested by \cite{Havei2015Journal}, who retrain the classification layer on patches extracted uniformly from the image. In practice, two stage training schemes can be prone to overfitting and sensitive to the state of the first classifier. Alternatively, \textit{dense training} (\cite{Long2014}) has been used to train a network on multiple or all voxels of a single image per optimisation step (\cite{urban2014CnnBrats, Brosch2015, Ronneberger2015}). This can introduce severe class imbalance, similarly to uniform sampling. Weighted cost functions have been proposed in the two latter works to alleviate this problem. \cite{Brosch2015} manually adjusted the sensitivity of the network, but the method can become difficult to calibrate for multi-class problems. \cite{Ronneberger2015} first balance the cost from each class, which has an effect similar to equal sampling, and further adjust it for the specific task by estimating the difficulty of segmenting each pixel.

\subsection{Contributions}

We present a fully automatic approach for lesion segmentation in multi-modal brain MRI based on an 11-layers deep, multi-scale, 3D CNN with the following main contributions:

\begin{enumerate}

\item We propose an efficient hybrid training scheme, utilizing \textit{dense training} (\cite{Long2014}) on sampled image segments, and analyze its behaviour in adapting to class imbalance of the segmentation problem at hand.

\item We analyze in depth the development of deeper, thus more discriminative, yet computationally efficient 3D CNNs. We exploit the utilization of small kernels, a design approach previously found beneficial in 2D networks (\cite{Simonyan2014}) that impacts 3D CNNs even more, and present adopted solutions that enable training deeper networks. 

\item We employ parallel convolutional pathways for multi-scale processing, a solution to efficiently incorporate both local and contextual information which greatly improves segmentation results.

\item We demonstrate the generalization capabilities of our system, which without significant modifications outperforms the state-of-the-art on a variety of challenging segmentation tasks, with top ranking results in two MICCAI challenges, ISLES and BRATS.

\end{enumerate}

Furthermore, a detailed analysis of the network reveals valuable insights into the powerful black box of deep learning with CNNs. For example, we have found that our network is capable of learning very complex, high level features that separate gray matter (GM), cerebrospinal fluid (CSF) and other anatomical structures to identify the image regions corresponding to lesions.

Additionally, we have extended the fully-connected Conditional Random Field (CRF) model by \cite{Krahenbuhl2013} to 3D which we use for final post-processing of the CNN's soft segmentation maps. This CRF overcomes limitations of previous models as it can handle arbitrarily large neighborhoods while preserving fast inference times. To the best of our knowledge, this is the first use of a fully connected CRF on medical data.

To facilitate further research and encourage other researchers to build upon our results, the source code of our lesion segmentation method including the CNN and the 3D fully connected CRF is made publicly available on \url{https://biomedia.doc.ic.ac.uk/software/deepmedic/}.


\section{Method}
\label{sec:segmentationSystem}

Our proposed lesion segmentation method consists of two main components, a 3D CNN that produces highly accurate, soft segmentation maps, and a fully connected 3D CRF that imposes regularization constraints on the CNN output and produces the final hard segmentation labels. The main contributions of our work are within the CNN component which we describe first in the following.

\subsection{3D CNNs for Dense Segmentation -- Setting the Baseline}
\label{subsec:theBaseline}

\begin{figure}[!h]
\centering
\iftrue
\begin{subfigure}[b]{1.0\textwidth}
\centering
	\includegraphics[clip=true, trim=0pt 0pt 0pt 0pt, width=1.0\textwidth]{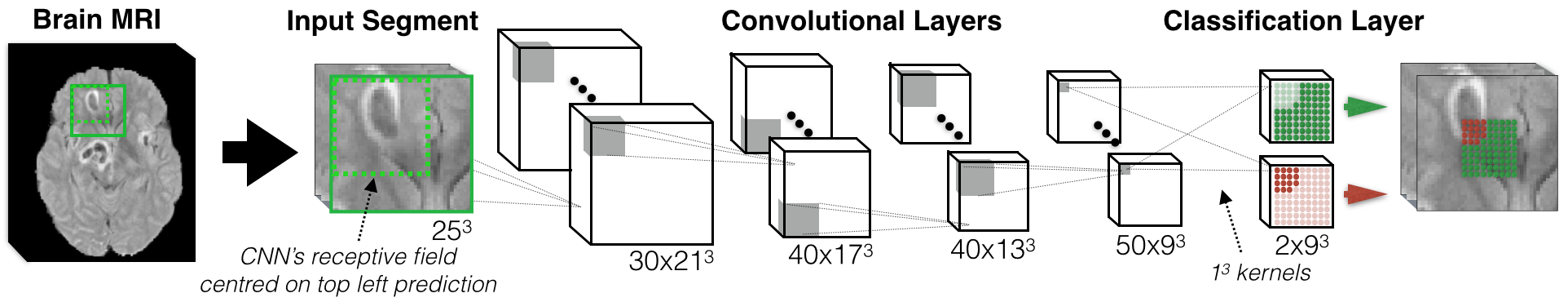}
\end{subfigure}
\fi
\caption{Our baseline CNN consists of four layers with $5^3$ kernels for feature extraction, leading to a receptive field of size $17^3$. The classification layer is implemented as convolutional with $1^3$ kernels, which enables efficient \textit{dense-inference}. When the network segments an input it predicts multiple voxels simultaneously, one for each shift of its receptive field over the input. Number of FMs and their size depicted as (\textit{Number $\times$ Size}).}
\label{fig:cnnBaseline}
\end{figure}

CNNs produce estimates for the voxel-wise segmentation labels by classifying each voxel in an image independently taking the neighborhood, i.e. local and contextual image information, into account. This is achieved by sequential convolutions of the input with multiple filters at the cascaded layers of the network. Each layer $l\in [1,L]$ consists of $C_l$ \textit{feature maps} (FMs), also referred to as \textit{channels}. Every FM is a group of neurons that detects a particular pattern, i.e. a feature, in the channels of the previous layer. The pattern is defined by the kernel weights associated with the FM. If the neurons of the $m$-th FM in the $l$-th layer are arranged in a 3D grid, their activations constitute the image $\mathbf{y}^{m}_{l} = f( \sum_{n=1}^{C_{l-1}}{\mathbf{k}^{m,n}_{l} \star \mathbf{y}^{n}_{l-1}} + b^{m}_{l})$. This is the result of convolving each of the previous layer's channels with a 3-dimensional \textit{kernel} $\mathbf{k}^{m,n}_{l}$, adding a learned bias $b^{m}_{l}$ and applying a non-linearity $f$. Each kernel is a matrix of learned hidden weights $\mathbf{W}^{m,n}_{l}$. The images $\mathbf{y}^n_0$, input to the first layer, correspond to the channels of the original input image, for instance a multi-sequence 3D MRI scan of the brain. The concatenation of the kernels $\mathbf{k}_l=(\mathbf{k}^{m,1}_{l},...,\mathbf{k}^{m,C_{l-1}}_{l})$ can be viewed as a 4-dimensional kernel convolving the concatenated channels $\mathbf{y}_{l-1}=(\mathbf{y}^{1}_{l-1}, ..., \mathbf{y}^{C_{l-1}}_{l-1})$, which then intuitively expresses that the neurons of higher layers combine the patterns extracted in previous layers, which results in the detection of increasingly more complex patterns. The activations of the neurons in the last layer $L$ correspond to particular segmentation class labels, hence this layer is also referred to as the classification layer. The neurons are thus grouped in $C_L$ FMs, one for each of the segmentation classes. Their activations are fed into a position-wise softmax function that produces the predicted posterior $p_c(\mathbf{x}) = \exp(\mathbf{y}_L^{c}(\mathbf{x}))/ \sum_{c=1}^{C_L} \exp(\mathbf{y}_L^{c}(\mathbf{x}))$ for each class $c$, which form soft segmentation maps with (pseudo-)probabilities. $\mathbf{y}_L^{c}(\mathbf{x})$ is the activation of the $c$-th classification FM at position $\mathbf{x} \in \mathbb{N}^3$. This baseline network is depicted in Fig.~\ref{fig:cnnBaseline}.

The neighborhood of voxels in the input that influence the activation of a neuron is its \textit{receptive field}. Its size, $\boldsymbol{\varphi}_l$, increases at each subsequent layer $l$ and is given by the 3-dimensional vector: 
\begin{equation} \label{eq:receptiveFile}
\boldsymbol{\varphi}_l^{\{ x,y,z\}} = \boldsymbol{\varphi}_{l-1}^{\{ x,y,z\}} + (\boldsymbol{\kappa}_l^{\{ x,y,z\}}-1) \boldsymbol{\tau}_l^{\{ x,y,z\}} \eqcm
\end{equation}
where $\boldsymbol{\kappa}_l,\boldsymbol{\tau}_l \in \mathbb{N}^3$ are vectors expressing the size of the kernels and stride of the receptive field at layer $l$. $\boldsymbol{\tau}_l$ is given by the product of the strides of kernels in layers preceding $l$. In this work only unary strides are used, as larger strides downsample the FMs (\cite{Springenberg2015}), which is unwanted behaviour for accurate segmentation. Thus in our system $\boldsymbol{\tau}_l = (1,1,1)$. The receptive field of a neuron in the classification layer corresponds to the image patch that influences the prediction for its central voxel. This is called the \textit{CNN's receptive field}, with $\boldsymbol{\varphi}_{CNN}=\boldsymbol{\varphi}_{L}$.

If input of size $\boldsymbol{\delta}_{in}$ is provided, the dimensions of the FMs in layer $l$ are given by:
\begin{equation} \label{eq:fmSize}
\boldsymbol{\delta}_l^{\{ x,y,z\}} = \lfloor (\boldsymbol{\delta}_{in}^{\{ x,y,z\}} - \boldsymbol{\varphi}_l^{\{ x,y,z\}}) / \boldsymbol{\tau}_l^{\{ x,y,z\}}	 +1 \rfloor
\end{equation}

In the common patch-wise classification setting, an input patch of size $\boldsymbol{\delta}_{in} = \boldsymbol{\varphi}_{CNN}$ is provided and the network outputs a single prediction for its central voxel. In this case the classification layer consists of FMs with size $1^3$. Networks that are implemented as fully-convolutionals are capable of \textit{dense-inference}, which is performed when input of size greater than $\boldsymbol{\varphi}_{CNN}$ is provided (\cite{Sermanet2013}). In this case, the dimensions of FMs increase according to Eq.~(\ref{eq:fmSize}). This includes the classification FMs which then output multiple predictions simultaneously, one for each stride of the CNN's receptive field on the input (Fig.~\ref{fig:cnnBaseline}). All predictions are equally trustworthy, as long as the receptive field is fully contained within the input and captures only original content, i.e. no padding is used. This strategy significantly reduces the computational costs and memory loads since the otherwise repeated computations of convolutions on the same voxels in overlapping patches are avoided. Optimal performance is achieved if the whole image is scanned in one forward pass. If GPU memory constraints do not allow it, such as in the case of large 3D networks where a large number of FMs need to be cached, the volume is tiled in multiple \textit{image-segments}, which are larger than individual patches, but small enough to fit into memory.

Before analyzing how we exploit the above dense-inference technique for training, which is the first main contribution of our work, we present the commonly used setting in which CNNs are trained patch-by-patch. Random patches of size $\boldsymbol{\varphi}_{CNN}$ are extracted from the training images. A \textit{batch} is formed out of $B$ of these samples, which is then processed by the network for one training iteration of Stochastic Gradient Descent (SGD). This step aims to alter the network's parameters $\mathbf{\Theta}$, such as weights and biases, in order to maximize the log likelihood of the data or, equally, minimize the Cross Entropy via the cost function:

\begin{equation} \label{eq:regCost}
J(\mathbf{\Theta}; \mathbf{I}^i, c^i)  = - \frac{1}{B} \sum_{i=1}^{B} \log\left(P(Y = c^i | \mathbf{I}^i, \mathbf{\Theta})\right) = - \frac{1}{B} \sum_{i=1}^{B} \log(p_{c^i}) \eqcm
\end{equation}
where the pair $(\mathbf{I}^i, c^i), \forall{i}\in{[1,B]}$ is the $i$-th patch in the batch and the true label of its central voxel, while the scalar value $p_{c^i}$ is the predicted posterior for class $c^i$. Regularization terms were omitted for simplicity. Multiple sequential optimization steps over different batches gradually lead to convergence.

\subsection{Dense Training on Image Segments and Class Balance}
\label{subsec:denseTraining}

Larger training batch sizes $B$ are preferred as they approximate the overall data more accurately and lead to better estimation of the true gradient by SGD. However, the memory requirement and computation time increase with the batch size. This limitation is especially relevant for 3D CNNs, where only a few dozens of patches can be processed within reasonable time on modern GPUs.

To overcome this problem, we devise a training strategy that exploits the dense inference technique on image segments. Following from Eq.~(\ref{eq:fmSize}), if an image segment of size greater than $\boldsymbol{\varphi}_{CNN}$ is given as input to our network, the output is a posterior probability for multiple voxels $V=\prod_{i=\{x,y,z\}}{\boldsymbol{\delta}_L^{(i)}}$. If the training batches are formed of $B$ segments extracted from the training images, the cost function (\ref{eq:regCost}) in the case of \textit{dense-training} becomes:

\begin{equation} \label{eq:costDense}
J_D(\mathbf{\Theta}; \mathbf{I}_s, \mathbf{c}_s) = - \frac{1}{B \cdot V} \sum_{s=1}^{B} \sum_{v=1}^{V} \log( p_{c_s^v}(\mathbf{x}^v)) \eqcm
\end{equation}

where $\mathbf{I}_s$ and $\mathbf{c}_s$ are the $s$-th segment of the batch and the true labels of its $V$ predicted voxels respectively. $c_s^v$ is the true label of the $v$-th voxel, $\mathbf{x}^v$ the corresponding position in the classification FMs and $p_{c_s^v}$ the output of the softmax function. The effective batch size is increased by a factor of $V$ without a corresponding increase in computational and memory requirements, as earlier discussed in Sec.~\ref{subsec:theBaseline}. Notice that this is a hybrid scheme between the commonly used training on individual patches and the dense training scheme on a whole image (\cite{Long2014}), with the latter being problematic to apply for training large 3D CNNs on volumes of high resolution due to memory limitations.


\begin{figure}[!h]
\centering
\begin{subfigure}[b]{0.5\textwidth}
\centering
	\includegraphics[clip=true, trim=0pt 0pt 0pt 0pt, width=1.0\textwidth]{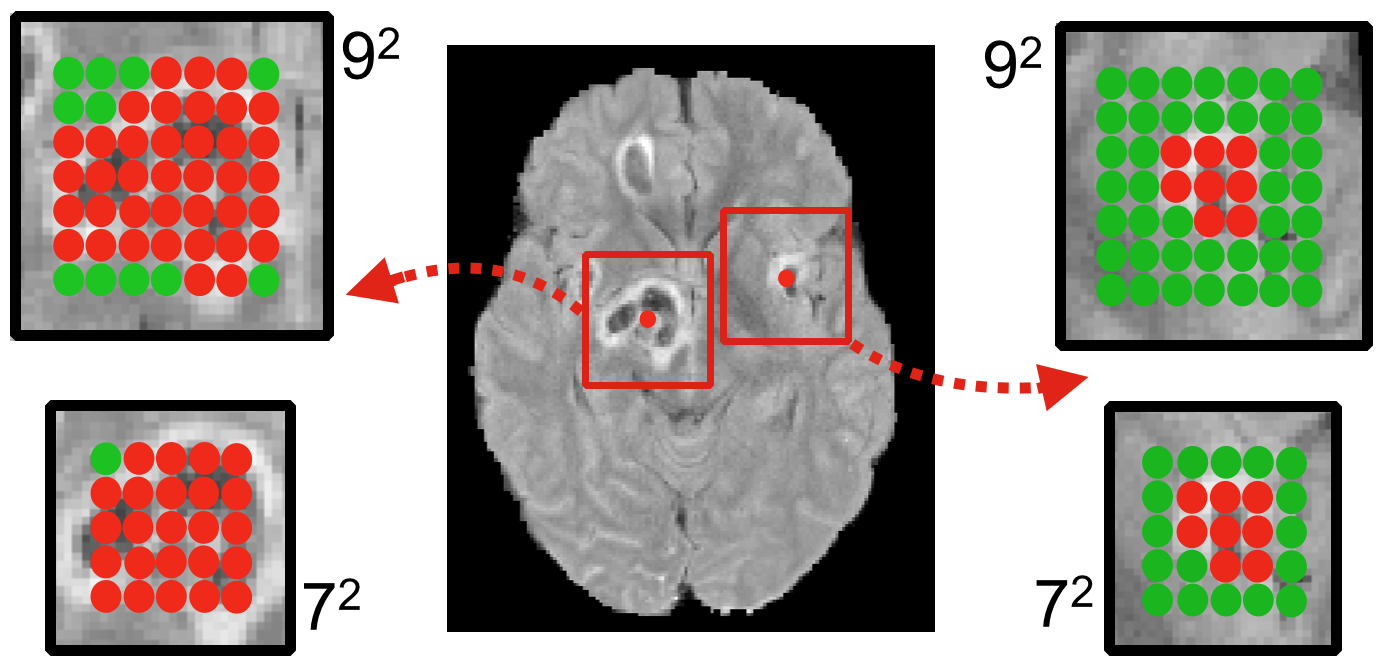}
\end{subfigure}

\caption{Consider a network with a 2D receptive field of $3^2$ (for illustration) densely-applied on the depicted lesion-centred image segments of size $7^2$ or $9^2$. Relatively more background (green) is captured by larger segments and around smaller lesions.}
\label{fig:segmentsVisual}
\end{figure}

An appealing consequence of this scheme is that the sampling of input segments provides a flexible and automatic way to balance the distribution of training samples from different segmentation classes which is an important issue that directly impacts the segmentation accuracy. Specifically, we build the training batches by extracting segments from the training images with 50\% probability being centred on a foreground or background voxel, alleviating class-imbalance. Note that the predicted voxels $V$ in a segment do not have to be of the same class, something that occurs when a segment is sampled from a region near class boundaries (Fig.~\ref{fig:segmentsVisual}). Hence, the sampling rate of the proposed hybrid method adjusts to the true distribution of the segmentation task's classes. Specifically, the smaller a labelled object, the more background voxels will be captured within segments centred on the foreground voxel. Implicitly, this yields a balance between sensitivity and specificity in the case of binary segmentation tasks. In multi-class problems, the rate at which different classes are captured within a segment centred on foreground reflects the real relative distribution of the foreground classes, while adjusting their frequency relatively to the background.

\subsection{Building Deeper Networks}
\label{subsec:buildingADeeperNetwork}
Deeper networks have greater discriminative power due to the additional non-linearities and better quality of local optima (\cite{Choromanska2015}). However, convolutions with 3D kernels are computationally expensive in comparison to the 2D variants, which hampers the addition of more layers. Additionally, 3D architectures have a larger number of trainable parameters, with each layer adding $C_l C_{l-1} \prod_{i=\{x,y,z\}}{\boldsymbol{\kappa}_{l}^{(i)}}$ weights to the model. $C_l$ is the number of FMs in layer $l$ and $\boldsymbol{\kappa}_{l}^{\{x,y,z\}}$ the size of its kernel in the respective spatial dimension. Overall this makes the network increasingly prone to over-fitting.

\begin{figure}[!h]
\centering
\begin{subfigure}[b]{0.5\textwidth}
\centering
	\includegraphics[clip=true, trim=0pt 0pt 0pt 0pt, width=1.0\textwidth]{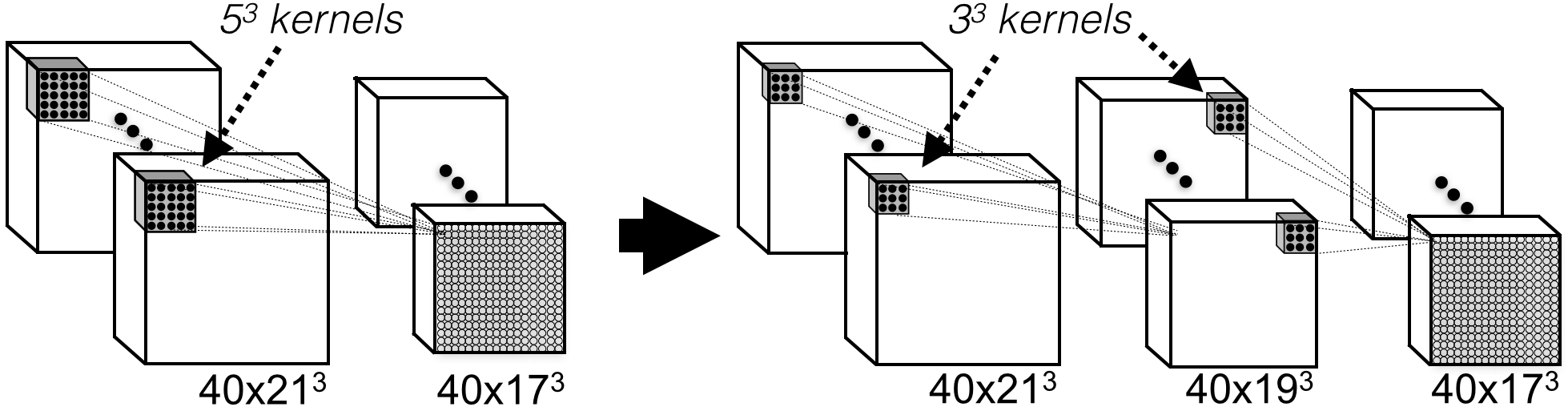}
\end{subfigure}

\caption{The replacement of the depicted layer with $5^5$ kernels (left) with two successive layers using $3^3$ kernels (right) introduces an additional non-linearity without altering the CNN's receptive field. Additionally, the number of weights is reduced from 200k to 86.4k and the required convolutions are cheaper (see text). Number of FMs and their size depicted as (\textit{Number $\times$ Size}).}
\label{fig:deeper3x3}
\end{figure}

In order to build a deeper 3D architecture, we adopt the sole use of small $3^3$ kernels that are faster to convolve with and contain less weights. This design approach was previously found beneficial for classification of natural images (\cite{Simonyan2014}) but its effect is even more drastic on 3D networks. When compared to common kernel choices of $5^3$ (\cite{zikic2014CnnBrats, urban2014CnnBrats, prasoon2013Knee}) and in our baseline CNN, the smaller $3^3$ kernels reduce the element-wise multiplications by a factor of approximately $5^3/3^3 \approx 4.6$ while reducing the number of trainable parameters by the same factor. Thus deeper network variants that are implicitly regularised and more efficient can be designed by simply replacing each layer of common architectures with more layers that use smaller kernels (Fig.~\ref{fig:deeper3x3}). 


However, deeper networks are more difficult to train. It has been shown that the forward (neuron activations) and backwards (gradients) propagated signal may explode or vanish if care is not given to retain its variance (\cite{Glorot2010}). This occurs because at every successive layer $l$, the variance of the signal is multiplied by $n^{in}_l \cdot var(\mathbf{W}_l)$, where $n^{in}_l=C_{l-1} \prod_{i=\{x,y,z\}}{\boldsymbol{\kappa}_{l}^{(i)}}$ is the number of weights through which a neuron of layer $l$ is connected to its input and $var(\mathbf{W}_l)$ is the variance of the layer's weights. To better preserve the signal in the initial training stage we adopt a scheme recently derived for ReLu-based networks by \cite{he2015delving} and initialize the kernel weights of our system by sampling from the normal distribution $\mathcal{N}(0,\sqrt{2/n^{in}_l})$.

A phenomenon of similar nature that hinders the network's performance is the \quot{internal covariate shift} (\cite{ioffe2015batch}). It occurs throughout training, because the weight updates to deeper layers result in a continuously changing distribution of signal at higher layers, which hinders the convergence of their weights. Specifically, at training iteration $t$ the weight updates may cause deviation $\epsilon_{l,t}$ to the variance of the weights. At the next iteration the signal will be amplified by $n^{in}_l \cdot var(\mathbf{W}_{l,t+1}) = n^{in}_l \cdot (var(\mathbf{W}_{l,t}) + \epsilon_{l,t})$. Thus before influencing the signal, any deviation $\epsilon_{l,t}$ is amplified by $n^{in}_l$ which is exponential in the number of dimensions. For this reason the problem affects training of 3D CNNs more severely than conventional 2D systems. For countering it, we adopt the recently proposed Batch Normalisation (BN) technique to all hidden layers (\cite{ioffe2015batch}), which allows normalization of the FM activations at every optimization step in order to better preserve the signal.

\subsection{Multi-Scale Processing via Parallel Convolutional Pathways}
\label{subsec:multiscaleCnn}

The segmentation of each voxel is performed by taking into account the contextual information that is captured by the receptive field of the CNN when it is centred on the voxel. The spatial context is providing important information for being able to discriminate voxels that otherwise appear very similar when considering only local appearance. From Eq.~(\ref{eq:receptiveFile}) follows that an increase of the CNN's receptive field requires bigger kernels or more convolutional layers, which increases computation and memory requirements. An alternative would be the use of pooling (\cite{LeCun1998}), which however leads to loss of the exact position of the segmented voxel and thus can negatively impact accuracy.

\begin{figure}[!h]
\centering
\begin{subfigure}[b]{1.0\textwidth}
\centering
	\includegraphics[clip=true, trim=0pt 0pt 0pt 0pt, width=1.0\textwidth]{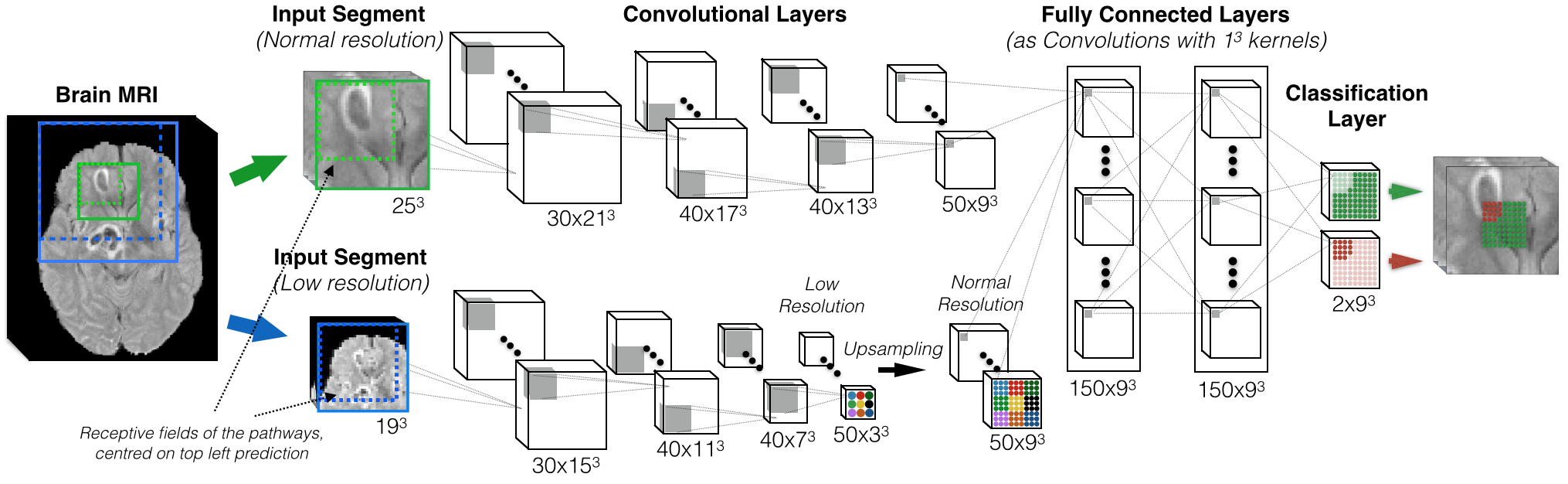}
\end{subfigure}
\caption{Multi-scale 3D CNN with two convolutional pathways. The kernels of the two pathways are here of size $5^3$ (for illustration only to reduce the number of layers in the figure). The neurons of the last layers of the two pathways thus have receptive fields of size $17^3$ voxels. The inputs of the two pathways are centered at the same image location, but the second segment is extracted from a down-sampled version of the image by a factor of 3. The second pathway processes context in an actual area of size $51^3$ voxels. \textit{DeepMedic}, our proposed 11-layers architecture, results by replacing each layer of the depicted pathways with two that use $3^3$ kernels (see Sec.~\ref{subsec:buildingADeeperNetwork}). Number of FMs and their size depicted as (\textit{Number $\times$ Size}).}
\label{fig:cnnMultiscale}
\end{figure}

In order to incorporate both local and larger contextual information into our 3D CNN, we add a second pathway that operates on down-sampled images. Thus, our dual pathway 3D CNN simultaneously processes the input image at multiple scales (Fig.~\ref{fig:cnnMultiscale}). Higher level features such as the location within the brain are learned in the second pathway, while the detailed local appearance of structures is captured in the first. As the two pathways are decoupled in this architecture, arbitrarily large context can be processed by the second pathway by simply adjusting the down-sampling factor $F_D$. The size of the pathways can be independently adjusted according to the computational capacity and the task at hand, which may require relatively more or less filters focused on the down-sampled context.


To preserve the capability of dense inference, spatial correspondence of the activations in the FMs of the last convolutional layers of the two pathways, $L1$ and $L2$, should be ensured. In networks  where only unary kernel strides are used, such as the proposed architecture, this requires that for every $F_D$ shifts of the receptive field $\boldsymbol{\varphi}_{L1}$ over the normal resolution input, only one shift is performed by $\boldsymbol{\varphi}_{L2}$ over the down-sampled input. Hence it is required that the dimensions of the FMs in $L2$ are $\boldsymbol{\delta}_{L2}^{\{ x,y,z\}} = \lceil \boldsymbol{\delta}_{L1}^{\{ x,y,z\}} / F_D \rceil$. From Eq.~(\ref{eq:fmSize}), the size of the input to the second pathway is $\boldsymbol{\delta}_{in2}^{\{ x,y,z\}} = \boldsymbol{\varphi}_{L2}^{\{ x,y,z\}} + \boldsymbol{\delta}_{L2}^{\{ x,y,z\}} -1 $ and similar is the relation between $\boldsymbol{\delta}_{in1}$ and $\boldsymbol{\delta}_{L1}$. These establish the relation between the required dimensions of the input segments from the two resolutions, which can then be extracted centered on the same image location. The FMs of $L2$ are up-sampled to match the dimensions of $L1$'s FMs and are then concatenated together. We add two more hidden layers for combining the multi-scale features before the final classification, as shown in Fig.~\ref{fig:cnnMultiscale}. Integration of the multi-scale parallel pathways in architectures with non-unary strides is discussed in \ref{app:detailsMultiscale}.

Combining multi-scale features has been found beneficial in other recent works (\cite{Long2014, Ronneberger2015}), in which whole 2D images are processed in the network by applying a few number of convolutions and then down-sampling the FMs for further processing at various scales. Our decoupled pathways allow arbitrarily large context to be provided while avoiding the need to load large parts of the 3D volume into memory. Additionally, our architecture extracts features completely independently from the multiple resolutions. This way, the features learned by the first pathway retain finest details, as they are not involved in processing low resolution context.

\subsection{3D Fully Connected CRF for Structured Prediction}
\label{subsec:3dFcCrf}

Because neighboring voxels share substantial spatial context, the soft segmentation maps produced by the CNN tend to be smooth, even though neighborhood dependencies are not modeled directly. However, local minima in training and noise in the input images can still result in some spurious outputs, with small isolated regions or holes in the predictions. We employ a fully connected CRF (\cite{Krahenbuhl2013}) as a post-processing step to achieve more structured predictions. As we describe below, this CRF is capable of modeling arbitrarily large voxel-neighborhoods but is also computationally efficient, making it ideal for processing 3D multi-modal medical scans.

For an input image $\mathbf{I}$ and the label configuration (segmentation) $\mathbf{z}$, the Gibbs energy in a CRF model is given by

\begin{equation}
E(\mathbf{z}) = \sum_i{\psi_u(z_i)} + \sum_{ij, i \neq j}{\psi_p(z_i,z_j)} \eqfs
\end{equation}

The unary potential is the negative log-likelihood $\psi_u(z_i) = -logP(z_i|\mathbf{I})$, where in our case $P(z_i|\mathbf{I})$ is the CNN's output for voxel $i$. In a fully connected CRF, the pairwise potential is of form $\psi_p(z_i,z_j) = \mu(z_i,z_j) k(\mathbf{f_i},\mathbf{f_j})$ between any pair of voxels, regardless of their spatial distance. The Pott's Model is commonly used as the label compatibility function, giving $\mu(z_i,z_j)=[z_i \neq z_j]$. The corresponding energy penalty is given by the function $k$, which is defined over an arbitrary feature space, with $\mathbf{f_i},\mathbf{f_j}$ being the feature vectors of the pair of voxels. \cite{Krahenbuhl2013} observed that if the penalty function is defined as a linear combination of Gaussian kernels, $k(\mathbf{f_i},\mathbf{f_j}) = \sum_{m=1}^{M} {w^{(m)}k^{(m)}(\mathbf{f_i},\mathbf{f_j})}$, the model lends itself for very efficient inference with mean field approximation, after expressing message passing as convolutions with the Gaussian kernels in the space of the feature vectors $\mathbf{f_i},\mathbf{f_j}$.

We extended the work of the original authors and implemented a 3D version of the CRF for processing multi-modal scans. We make use of two Gaussian kernels, which operate in the feature space defined by the voxel coordinates $p_{i,d}$ and the intensities of the $c$-th modality-channel $I_{i,c}$ for voxel $i$. The smoothness kernel, $k^{(1)}(\mathbf{f_i}, \mathbf{f_j}) = exp\Big(- \sum_{d=\{x,y,z\}}{ \frac{\vert p_{i,d} - p_{j,d} \vert ^2}{2\sigma_{\alpha, d}^2} } \Big)$, is defined by a diagonal covariance matrix with elements the configurable parameters $\sigma_{\alpha, d}$, one for each axis. These parameters express the size and shape of neighborhoods that homogeneous labels are encouraged. The appearance kernel $ k^{(2)}(\mathbf{f_i},\mathbf{f_j}) = exp \Big( - \sum_{d=\{x,y,z\}}{ \frac{\vert p_{i,d}-p_{j,d} \vert ^2}{2\sigma_{\beta,d}^2} } - \sum_{c=1}^{C}{ \frac{\vert I_{i,c} - I_{j,c} \vert ^2}{2\sigma_{\gamma,c}^2} } \Big) $ is defined similarly. The additional parameters $\sigma_{\gamma,c}$ can be interpreted as how strongly to enforce homogeneous appearance in the $C$ input channels, when voxels in an area spatially defined by $\sigma_{\beta,d}$ are identically labelled. Finally, the configurable weights $w^{(1)},w^{(2)}$ define the relative strength of the two factors.



\section{Analysis of Network Architecture}
\label{sec:vaOfNetArch}

In this section we present a series of experiments in order to analyze the impact of each of the main contributions and to justify the choices made in the design of the proposed 11-layers, multi-scale 3D CNN architecture, referred to as the \textit{DeepMedic}. Starting from the CNN baseline as discussed in Sec.~\ref{subsec:theBaseline}, we first explore the benefit of our proposed dense training scheme (cf. Sec.~\ref{subsec:denseTraining}), then investigate the use of deeper models (cf. Sec.~\ref{subsec:buildingADeeperNetwork}) and then evaluate the influence of the multi-scale dual pathway (cf. Sec.~\ref{subsec:multiscaleCnn}). Finally, we compare our method with corresponding 2D variants to assess the benefit of processing 3D context.

\subsection{Experimental Setting}
\label{subsec:experimentSetting}

The following experiments are conducted using the TBI dataset with 61 multi-channel MRIs which is described in more detail later in Sec.~\ref{subsec:evalTbi}. Here, the images are randomly split into a validation and training set, with 15 and 46 images each. The same sets are used in all analyses. To monitor the progress of segmentation accuracy during training, we extract 10k random patches at regular intervals, with equal numbers extracted from each of the validation images. The patches are uniformly sampled from the brain region in order to approximate the true distribution of lesions and healthy tissue. Full segmentation of the validation datasets is performed every five epochs and the mean Dice similarity coefficient (DSC) is determined. Details on the configuration of the networks are provided in \ref{app:detailsConfig}.

\subsection{Effect of Dense Training on Image Segments}
\label{subsec:valDenseTraining}

\begin{figure}[!h]
\centering
\begin{subfigure}[b]{1.0\textwidth}
\centering
	\includegraphics[clip=true, trim=0pt 0pt 0pt 0pt, width=1.0\textwidth]{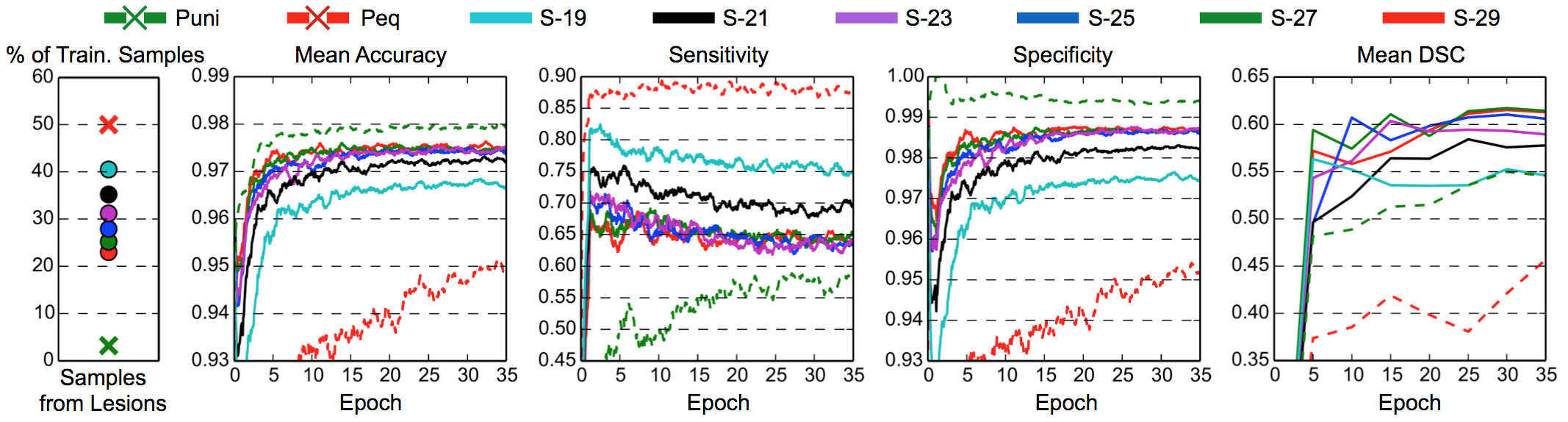}
\end{subfigure}
\caption{Comparison of the commonly used methods for training on patches uniformly sampled from the brain region (P$_\text{uni}$) and equally sampled from lesion and background (P$_\text{eq}$) against our proposed scheme (S-${d}$) on cubic segments of side length $d$, also equally sampled from lesion and background. We varied $d$ to observe its effect. From left to right: percentage of training samples extracted from the lesion class, mean accuracy, sensitivity, specificity calculated on uniformly sampled validation patches and, finally, the mean DSC of the segmentation of the validation datasets. The progress throughout training is plotted. Because lesions are small, P$_\text{uni}$ achieves very high voxel-wise accuracy by being very specific but not sensitive, with the opposite being the case for P$_\text{eq}$. Our method achieves an effective balance between the two, resulting in better segmentation as reflected by higher DSC.}
\label{fig:denseTrainingExperiment}
\end{figure}

We compare our proposed dense training method with two other commonly used training schemes on the 5-layers baseline CNN (see Fig.~\ref{fig:cnnBaseline}). The first common scheme trains on $17^3$ patches extracted uniformly from the brain region, and the second scheme samples patches equally from the lesion and background class. We refer to these schemes as P$_\text{uni}$ and P$_\text{eq}$. The results shown in Fig.~\ref{fig:denseTrainingExperiment} show a correlation of sensitivity and specificity with the percentage of training samples that come from the lesion class. P$_\text{eq}$ performs poorly because of over-segmentation (high sensitivity, low specificity). P$_\text{uni}$ has better classification on the background class (high specificity), which leads to high mean voxel-wise accuracy since the majority corresponds to background, but not particularly high DSC scores due to under-segmentation (low sensitivity).

To evaluate our dense training scheme, we train multiple models with varying sized image segments, equally sampled from lesions and background. The tested sizes of the segments go from $19^3$ upwards to $29^3$. The models are referred to as \quot{S-$d$}, where $d$ is the side length of the cubic segments. For fair comparison, the batch sizes in all the experiments are adjusted to have a similar memory footprint and lead to similar training times as compared to training on P$_{uni}$ and P$_{eq}$\footnote{Dense training on a whole volume was inapplicable in these experimental settings due to memory limitations but was previously shown to give similar results as training on uniformly sampled patches (\cite{Long2014}).}. We observe a great performance increase for model S-${19}$ over P$_\text{eq}$. We account this partly to the efficient increase of the effective batch size ($B \cdot V$ in Eq.~(\ref{eq:costDense})), but also to the altered distribution of training samples. As we increase the size of the training segments further, we quickly reach a balance between the sensitivity of P$_{eq}$ and the specificity of P$_{uni}$, which results in improved segmentation as expressed by the DSC.

The segment size is a hyper-parameter in our model. We observe that the increase in performance with increasing segment size quickly levels off, and similar performance is obtained for a wide range of segment sizes, which allows for easy configuration. For the remaining experiments, all models were trained on segments of size $25^3$.

\subsection{Effect of Deeper Networks}
\label{subsec:valDeeper}

\begin{figure}[!h]
\centering
\begin{subfigure}[b]{0.5\textwidth}
\centering
	\includegraphics[clip=true, trim=0pt 0pt 0pt 0pt, width=1.0\textwidth]{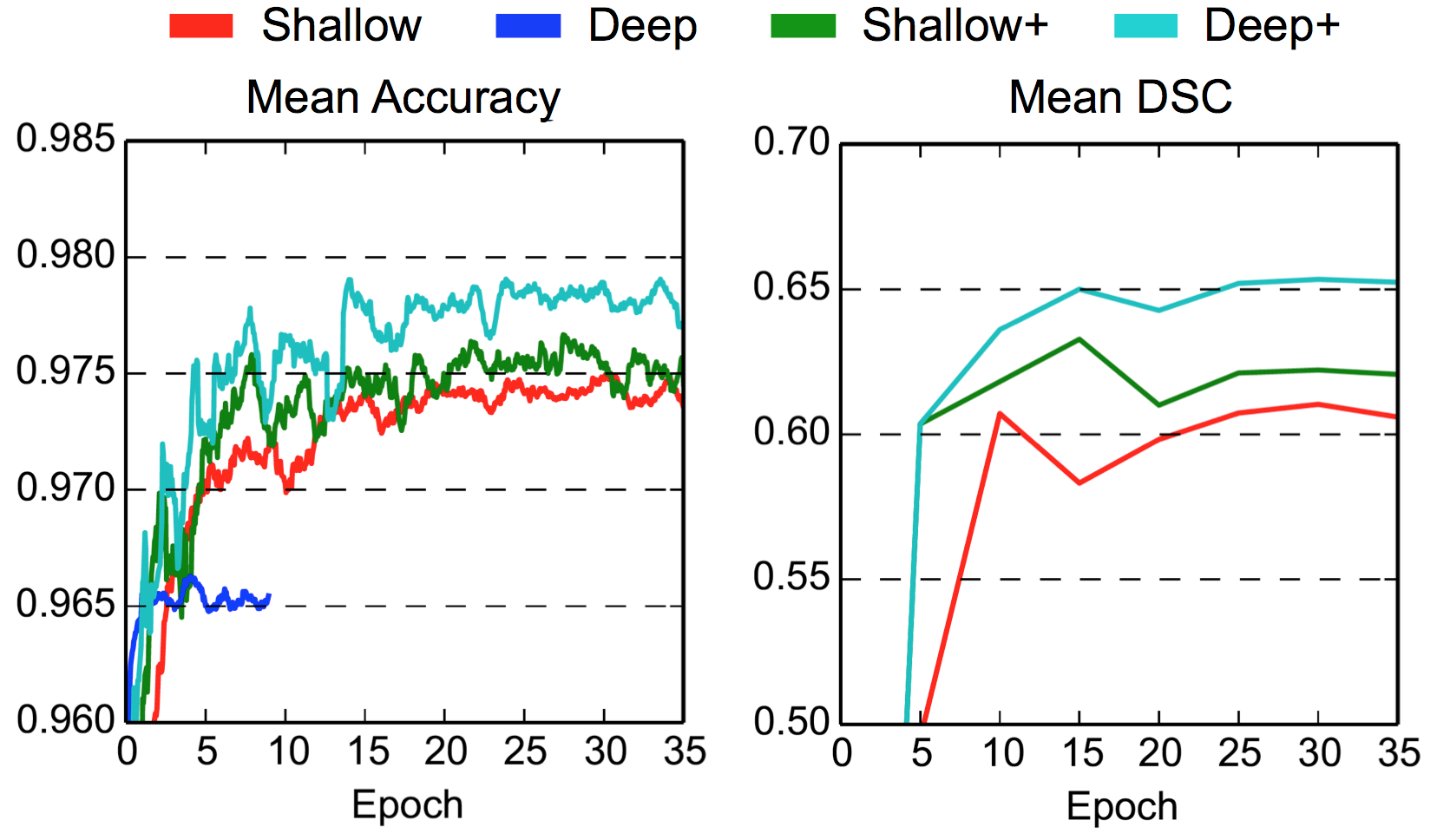}
\end{subfigure}
\caption{Mean accuracy over validation samples and DSC for the segmentations of the validation images, as obtained from the \quot{Shallow} baseline and \quot{Deep} variant with smaller kernels. Training of the plain deeper model fails (cf. Sec.~\ref{subsec:valDeeper}). This is overcome by adopting the initialization scheme of (\cite{he2015delving}), which further combined with Batch Normalization leads to the enhanced (\texttt{+}) variants. Deep\texttt{+} performs significantly better than Shallow\texttt{+} with similar computation time, thanks to the use of small kernels.
}
\label{fig:deepProblems}
\end{figure}

The 5-layers baseline CNN (Fig.~\ref{fig:cnnBaseline}), here referred to as the \quot{Shallow} model, is extended to 9-layers by replacing each convolutional layer that uses $5^3$ kernels with two layers that use $3^3$ kernels (Fig.~\ref{fig:deeper3x3}). This model is referred to as \quot{Deep}. Training the latter, however, utterly fails with the model making only predictions corresponding to the background class. This problem is related to the challenge of preserving the signal as it propagates through deep networks and its variance gets multiplied with the variance of the weights, as previously discussed in Sec.~\ref{subsec:buildingADeeperNetwork}. One of the causes is that the weights of both models have been initialized with the commonly used scheme of sampling from the normal distribution $\mathcal{N}(0,0.01)$ (cf. \cite{Krizhevsky2012}). In comparison, the initialization scheme by \cite{he2015delving}, derived for preserving the signal in the initial stage of training, results in higher values and overcomes this problem. Further preservation of the signal is obtained by employing Batch Normalization. This results in an enhanced 9-layers model which we refer to as \quot{Deep\texttt{+}}, and using the same enhancements on the Shallow model yields \quot{Shallow\texttt{+}}. The significant performance improvement of Deep\texttt{+} over Shallow\texttt{+}, as shown in Fig.~\ref{fig:deepProblems}, is the result of the greater representational power of the deeper network. The two models need similar computational times, which highlights the benefits of utilizing small kernels in the design of 3D CNNs. Although the deeper model requires more sequential (layer by layer) computations on the GPU, those are faster due to the smaller kernel size.

\subsection{Effect of the Multi-Scale Dual Pathway}
\label{subsec:valMultiscale}

\begin{figure}[!h]
\centering
\begin{subfigure}[b]{0.5\textwidth}
\centering
	\includegraphics[clip=true, trim=0pt 0pt 0pt 0pt, width=1.0\textwidth]{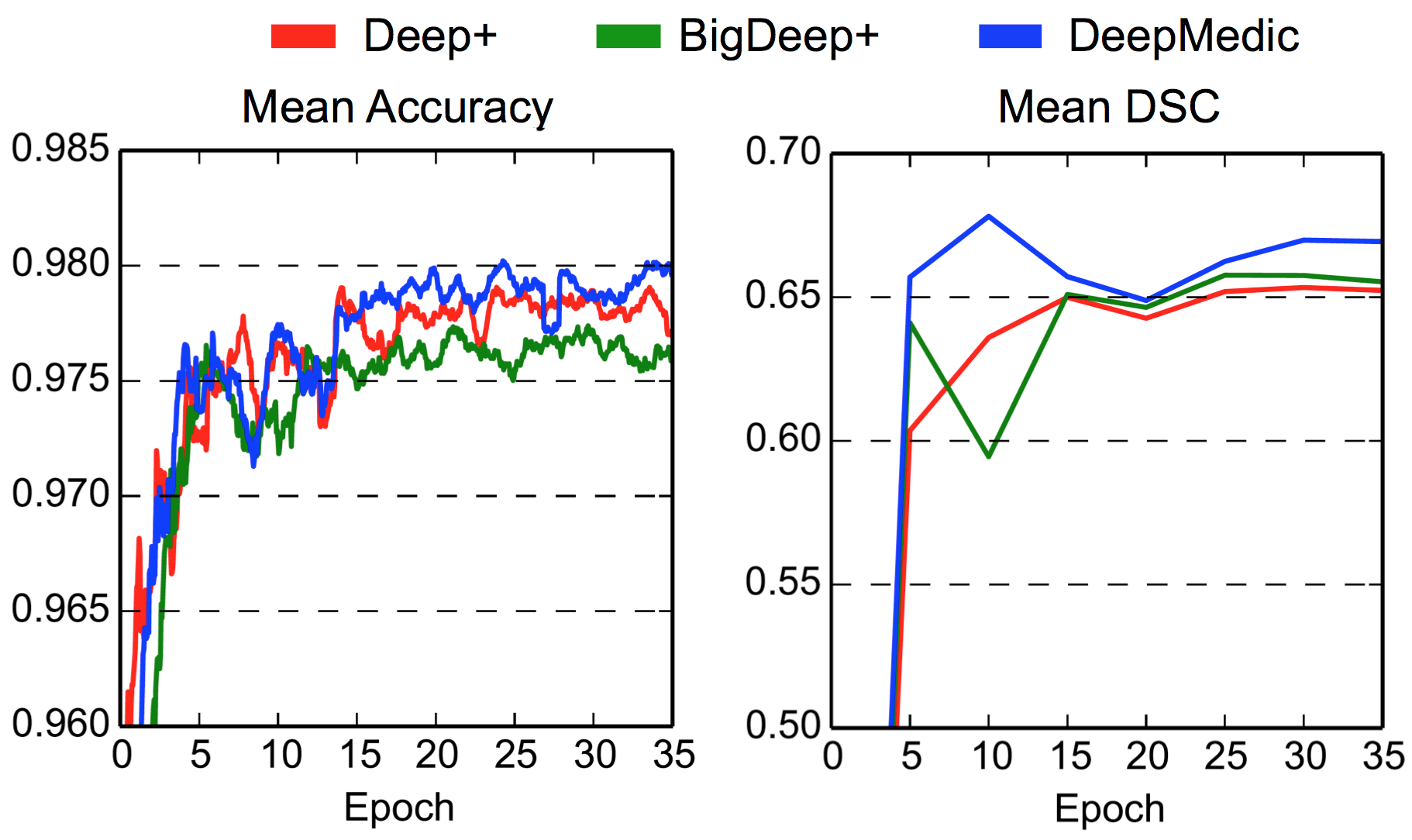}
\end{subfigure}
\caption{Mean accuracy over validation samples and DSC for the segmentation of the validation images, as obtained by a single-scale model (Deep\texttt{+}) and our dual pathway architecture (DeepMedic). We also trained a single-scale model with larger capacity (BigDeep\texttt{+}), similar to the capacity of DeepMedic. DeepMedic yields best performance by capturing greater context, while BigDeep\texttt{+} seems to suffer from over-fitting.
}
\label{fig:multiscaleExperiment}
\end{figure}

The final version of the proposed network architecture, referred to as \quot{DeepMedic}, is built by extending the Deep\texttt{+} model with a second convolutional pathway that is identical to the first one. Two hidden layers are added for combining the multi-scale features before the classification layer, resulting in a deep network of 11-layers (cf. Fig.~\ref{fig:cnnMultiscale}). The input segments to the second pathway are extracted from the images down-sampled by a factor of three. Thus, the network is capable of capturing context in a $51^3$ area of the original image through the $17^3$ receptive field of the lower-resolution pathway, while only doubling the computational and memory requirements over the single pathway CNN. In comparison, the most recent 2D CNN systems proposed for lesion segmentation (\cite{Havei2015Journal, pereira2015Brats}) have a receptive field limited to $33^2$ voxels.

\begin{figure}[!h]
\centering
\begin{subfigure}[b]{0.85\textwidth}
\centering
	\includegraphics[clip=true, trim=0pt 0pt 0pt 0pt, width=1.0\textwidth]{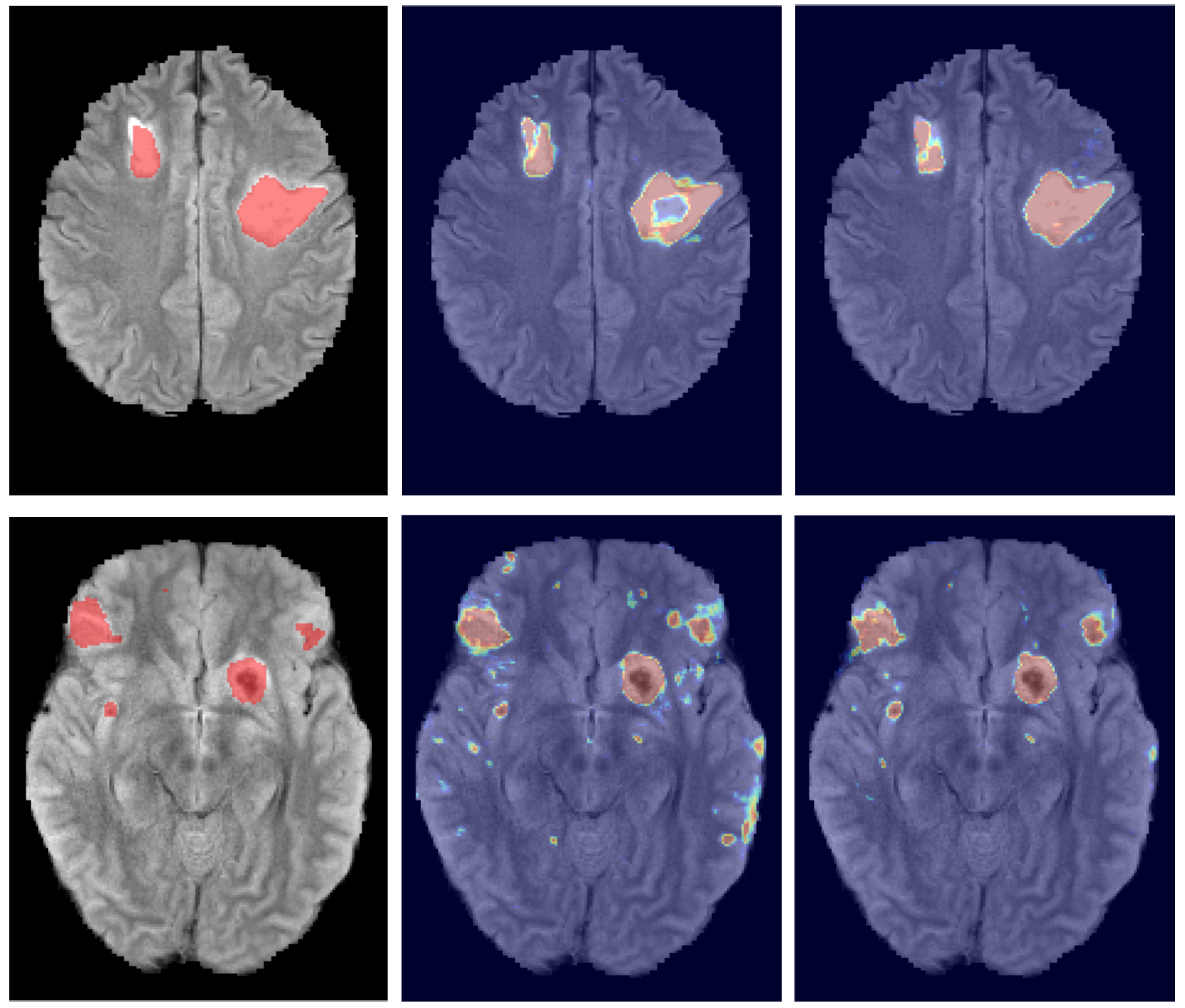}
\end{subfigure}

\caption{(Rows) Two cases from the severe TBI dataset, showing representative improvements when using the multi-scale CNN approach. (Columns) From left to right: the MRI FLAIR sequence with the manually labeled lesions, predicted soft segmentation map obtained from a single-scale model (Deep\texttt{+}) and the prediction of the multi-scale DeepMedic model. The incorporation of greater context enables DeepMedic to identify when it processes an area within larger lesions (top). Spurious false positives are significantly reduced across the image on the bottom.}
\label{fig:qualitativeMultiscaleVal}
\end{figure}

Figure~\ref{fig:multiscaleExperiment} shows the improvement DeepMedic achieves over the single pathway model Deep\texttt{+}. In Fig.~\ref{fig:qualitativeMultiscaleVal} we show two representative visual examples of this improvement when using the multi-scale CNN. Finally, we confirm that the performance increase can be accounted to the additional context and not the additional capacity of DeepMedic. To this end, we build a big single-scale model by doubling the FMs at each of the 9-layers of Deep\texttt{+} and adding two hidden layers. This 11-layers deep and wide model, referred to as \quot{BigDeep\texttt{+}}, has the same number of parameters as DeepMedic. The performance of the model is not improved, while showing signs of over-fitting.

\subsection{Processing 3D in comparison to 2D Context}
\label{subsec:val3dContext}

Acquired brain MRI scans are often anisotropic. Such is the case for most sequences in our TBI dataset, which have been acquired with lower axial resolution, except for the isotropic MPRAGE. We perform a series of experiments to investigate the behaviour of 2D networks and assess the benefit of processing 3D context in this setting.

DeepMedic can be converted to 2D by setting the third dimension of each kernel to one. This way only information from the surrounding context on the axial plane influences the classification of each voxel. If 2D segments are given as input, the dimensionality of the feature maps decreases and so does the memory required. This allows developing 2D variants with increased width, depth and size of training batch with similar requirements as the 3D version, which are valid candidates for model selection in practical scenarios. We assess various configurations and present some representatives in Table \ref{subtab:netsConfig2d} along with their performance. Best segmentation among investigated 2D variants is achieved by a 19-layers, multi-scale network, reaching 61.5\% average DSC on the validation fold. The decline from the 66.6\% DSC achieved by the 3D version of DeepMedic indicates the importance of processing 3D context even in settings where most acquired sequences have low resolution along a certain axis.



\section{Evaluation on Clinical Data}
\label{sec:evaluation}

The proposed system consisting of the DeepMedic CNN architecture, optionally coupled with a fully connected CRF, is evaluated on three lesion segmentation tasks including challenging clinical data from patients with traumatic brain injuries, brain tumors, and ischemic stroke. Quantitative evaluation and comparisons with state-of-the-art are reported for each of the tasks.

\subsection{Traumatic Brain Injuries}
\label{subsec:evalTbi}

\subsubsection{Material and Pre-Processing}
\label{subsubsec:materialTbi}

Sixty-six patients  with moderate-to-severe TBI who required admission to the Neurosciences Critical Care Unit at Addenbrooke's Hospital, Cambridge, UK, underwent imaging using a 3-Tesla Siemens Magnetom TIM Trio within the first week of injury. Ethical approval was obtained from the Local Research Ethics Committee (LREC 97/290) and written assent via consultee agreement was obtained for all patients. The structural MRI sequences that are used in this work are isotropic MPRAGE (1$mm\times$1$mm\times$1$mm$), axial FLAIR, T2 and Proton Density (PD) (0.7$mm\times$0.7$mm\times$5$mm$), and Gradient-Echo (GE) (0.86$mm\times$0.86$mm\times$5$mm$). All visible lesions were manually annotated on the FLAIR and GE sequences with separate labeling for each lesion type. In nine patients the presence of hyperintense white matter lesions that were felt to be chronic in nature were also annotated. Artifacts, for example, signal loss secondary to intraparenchymal pressure probes, were also noted. For the purpose of this study we focus on binary segmentation of all abnormalities within the brain tissue. Thus, we merged all classes that correspond to intra-cerebral abnormalities into a single \quot{lesion} label. Extra-cerebral pathologies such as epidural and subdural hematoma were treated as background. We excluded two datasets because of corrupted FLAIR images\ignore{13296, 17792}, two cases because no lesions were found\ignore{13776, 15883} and one case \ignore{11976} because of a major scanning artifact corrupting the images. This results in a total of 61 cases used for quantitative evaluation. Brain masks were obtained using the ROBEX tool (\cite{Iglesias2011}). All images were resampled to an isotropic $1mm^3$ resolution, with dimensions 193$\times$229$\times$193 and affinely registered (\cite{Studholme1999}) to MNI space using the atlas by \cite{Grabner2006}. No bias field correction was used as preliminary results showed that this can negatively affect lesion appearance. Image intensities were normalized to have zero-mean and unit variance, as it has been reported that this improves CNN results (\cite{Jarrett2009}).

\subsubsection{Experimental Setting}
\label{subsubsec:tbiExperimentalSetting}

\textbf{Network configuration and training:} The network architecture corresponds to the one described in Sec.~\ref{subsec:valMultiscale}, i.e. a dual-pathway, 11-layers deep CNN. The training data is augmented by adding images reflected along the sagittal axis. To make the network invariant to absolute intensities we also shift the intensities of each MR channel $c$ of every training segment by $i_c = r_c \sigma_c$. $r_c$ is sampled for every segment from $\mathcal{N}(0,0.1)$ and $\sigma_c$ is the standard deviation of intensities under the brain mask in the corresponding image. The network is regularized using dropout (\cite{hinton2012dropout}) with a rate of 2\% on all convolutional layers, which is in addition to a 50\% rate used on the last two layers. The network is evaluated with 5-fold cross-validation on the 61 subjects.

\textbf{CRF configuration:} The parameters of the fully connected CRF are determined in a configuration experiment using random-search and 15 randomly selected subjects from the TBI database with predictions from a preliminary version of the corresponding model. The 15 subjects are reshuffled into the 5-folds used for subsequent evaluation.

\textbf{Random Forest baseline:} We have done our best to set up a competitive baseline for comparison. We employ a context-sensitive Random Forest, similar to the model presented by \cite{Zikic2012} for brain tumors except that we apply the forest to the MR images without additional tissue specific priors. We train a forest with 50 trees and maximum depth of 30. Larger size did not improve results. Training data points are approximately equally sampled from lesion and background classes, with the optimal balance empirically chosen. Two hundred randomized cross-channel box features are evaluated at each split node with maximum offsets and box sizes of 20mm. The same folds of training and test sets are used as for our CNN approach.

\subsubsection{Results}
\label{subsec:resTbi}

\begin{table}[!h]
\centering
\scriptsize
\caption{Performance of \textit{DeepMedic} and an \textit{ensemble} of three networks on the TBI database. For comparison, we provide results for a Random Forest baseline. Values correspond to the mean (and standard deviation). Numbers in bold indicate significant improvement by the CRF post-processing, according to a two-sided, paired t-test on the DSC metric (*$p<5 \cdot 10^{-2}$, **$p<10^{-4}$).}
\label{table:accuracyTbiTrio}
\begin{tabular}{@{}llllll@{}}
\toprule
\multicolumn{1}{c}{}	& DSC			& Precision		& Sensitivity		& ASSD					& Haussdorf 	\\ \midrule
R. Forest			& 51.1(20.0)		& 50.1(24.4) 	& 60.1(15.8)			& 8.29(6.76)				& 64.17(15.98)	\\
R. Forest+CRF		& \textbf{54.8(18.5)**}	& 58.6(23.1)	& 56.9(17.4)		& 6.71(5.01)				& 59.45(15.52)	\\
DeepMedic			& 62.3(16.4)		& 65.3(18.8)		& 64.4(16.3)			& 4.24(2.64)				& 56.50(15.88)	\\
DeepMedic+CRF		& \textbf{63.0(16.3)**} & 67.7(18.2)	& 63.2(16.7)		& 4.02(2.54)				& 55.68(15.93)	\\
Ensemble				& 64.2(16.2)		& 67.7(18.3)		& 65.3(16.3)			& 3.88(2.33)				& 54.38(15.45)	\\
Ensemble+CRF			& \textbf{64.5(16.3)*} 	& 69.8(17.8)		& 63.9(16.7)		& 3.72(2.29)				&52.38(16.03)	\\
\bottomrule
\end{tabular}
\end{table}

\begin{figure}[!ht]
\centering
\begin{subfigure}[b]{1.0\textwidth}
	\centering
	\includegraphics[clip=true, trim=0pt 0pt 0pt 0pt, width=1.0\textwidth]{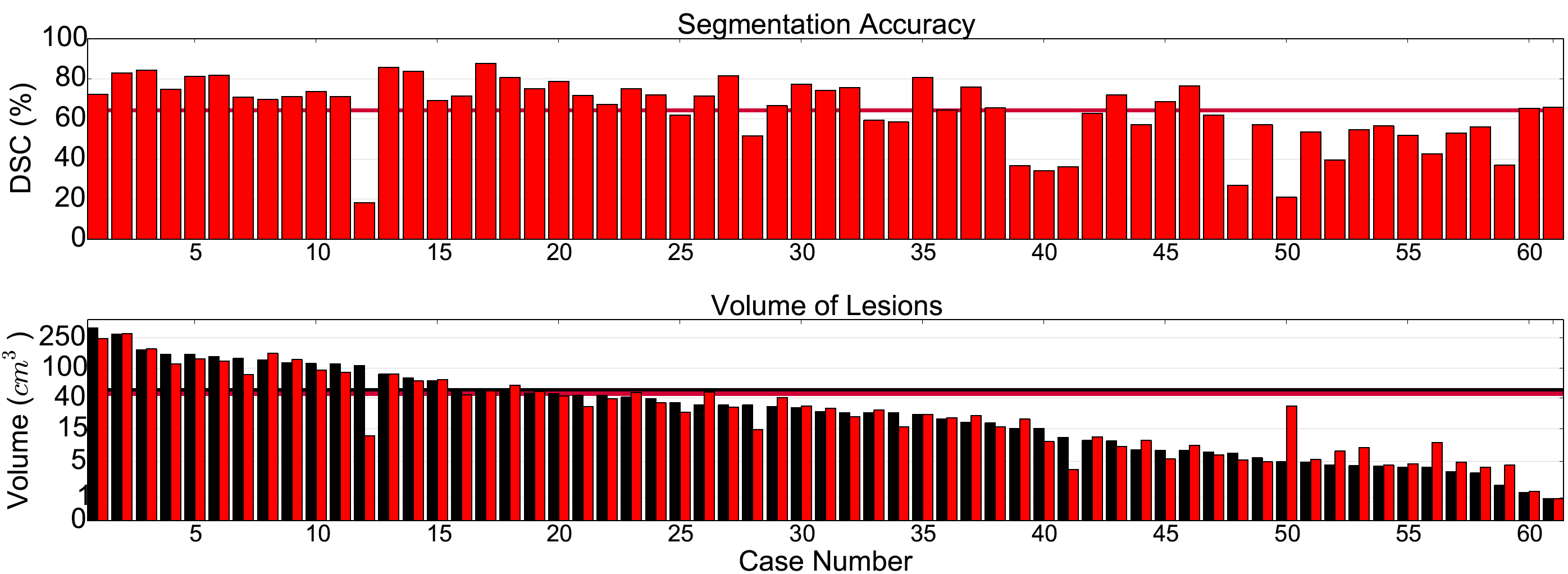}
\end{subfigure}
\vspace{-0pt} 
\caption{(Top) DSC achieved by our ensemble of three networks on each of the 61 TBI datasets. (Bottom) Manually segmented (black) and predicted lesion volumes (red). Note here the logarithmic scale. Continuous lines represent mean values. The outlying subject 12 presents small TBI lesions, which are successfully segmented, but also vascular ischemia. Because it is the only case in the database with the latter pathology, the networks fail to segment it as such lesion was not seen during training.}
\label{fig:evalTbiAccVsVol}
\vspace{-10pt}
\end{figure}

\begin{figure}[!ht]
\centering
\begin{subfigure}[b]{1.0\textwidth}
	\centering
	\includegraphics[clip=true, trim=0pt 0pt 0pt 0pt, width=1.0\textwidth]{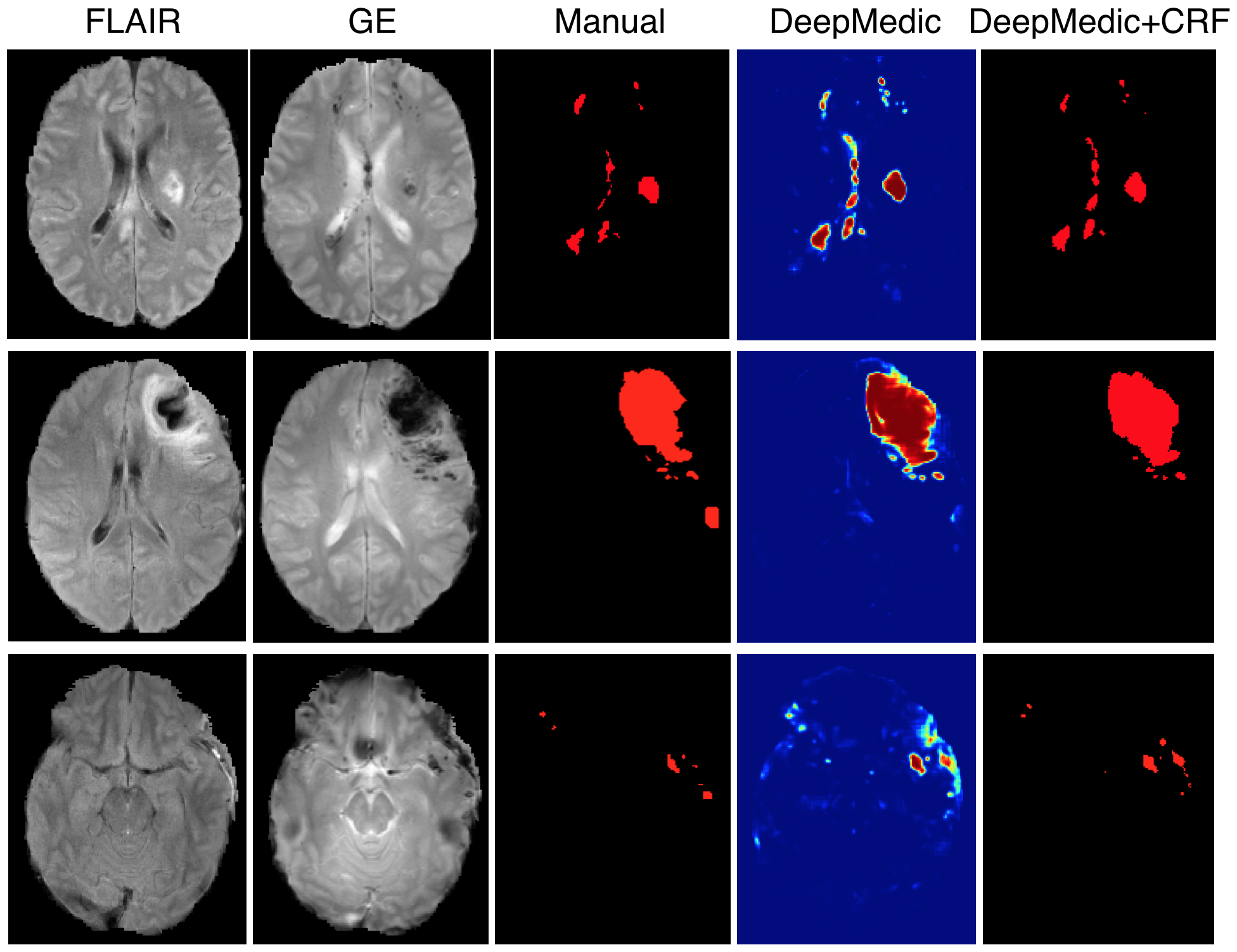}
\end{subfigure}
\vspace{-10pt} 
\caption{Three examples from the application of our system on the TBI database. It is capable of precise segmentation of both small and large lesions. Second row depicts one of the common mistakes observed. A contusion near the edge of the brain is under-segmented, possibly mistaken for background. Bottom row shows one of the worst cases, representative of the challenges in segmenting TBI. Post-surgical sub-dural debris is mistakenly captured by the brain mask. The network partly segments the abnormality, which is not a celebral lesion of interest.}
\label{fig:evalTbiVisualQuality}
\end{figure}

Table \ref{table:accuracyTbiTrio} summarizes the results on TBI. Our CNN significantly outperforms the Random Forest baseline, while the relatively overall low DSC values indicate the difficulty of the task.  Due to randomness during training the local minima where a network converges are different between training sessions and some errors they produce differ (\cite{Choromanska2015}). To clear the unbiased errors of the network we form an \textit{ensemble} of three similar networks, aggregating their output by averaging. This ensemble yields better performance in all metrics but also allows us to investigate the behaviour of our network focusing only on the biased errors. Fig.~\ref{fig:evalTbiAccVsVol} shows the DSC obtained by the ensemble on each subject in relation to the manually segmented and predicted lesion volume. The network is capable of segmenting cases with very small lesions, although, performance is less robust in these cases as even small errors have large influence on the DSC metric. Investigation of the predicted lesion volume, which is an important biomarker for prognostication, shows that the network is neither biased towards the lesion nor background class, with promising results even on cases with very small lesions. Furthermore, we separately evaluate the influence of the post-processing with the fully connected CRF. As shown in Table \ref{table:accuracyTbiTrio}, the CRF yields improvements over all classifiers. Effects are more prominent when the performance of the primary segmenter degrades, which shows the robustness of this regulariser. Fig.~\ref{fig:evalTbiVisualQuality} shows three representative cases.


\subsection{Brain Tumor Segmentation}
\label{subsec:evalBrats}

\subsubsection{Material and Pre-Processing}

For brain tumors, we evaluate our system on the data from the 2015 Brain Tumor Segmentation Challenge (BRATS) (\cite{Menze2014}). The training set consists of 220 cases with high grade (HG) and 54 cases with low grade (LG) glioma for which corresponding reference segmentations are provided. The segmentations include the following tumor tissue classes: 1) necrotic core, 2) edema, 3) non-enhancing and 4) enhancing core. The test set consists of 110 cases of both HG and LG but the grade is not revealed. Reference segmentations for the test set are hidden and evaluation is carried out via an online system. For evaluation, the four predicted labels are merged into different sets of whole tumor (all four classes), the core (classes 1,3,4), and the enhancing tumor (class 4)\footnote{For interpretation of the results note that, to the best of our knowledge, cases where the \quot{enhancing tumor} class is not present in the manual segmentation are considered as zeros for the calculation of average performance by the evaluation platform, lowering the upper bound for this class.}. For each subject, four MRI sequences are available, FLAIR, T1, T1-contrast and T2. The datasets are pre-processed by the organizers and provided as skull-stripped, registered to a common space and resampled to isotropic $1mm^3$ resolution. Dimensions of each volume are 240$\times$240$\times$155. We add minimal pre-processing of normalizing the brain-tissue intensities of each sequence to have zero-mean and unit variance.

\subsubsection{Experimental Setting}

\textbf{Network configuration and training:} We modify the DeepMedic architecture to handle multi-class problems by extending the classification layer to five feature maps (four tumor classes plus background). The rest of the configuration remains unchanged. We enrich the dataset with sagittal reflections. Opposite to the experiments on TBI, we do not employ the intensity perturbation and dropout on convolutional layers, because the network should not require as much regularisation with this large database. The network is trained on image segments extracted with equal probability centred on the whole tumor and healthy tissue. The distribution of the classes captured by our training scheme is provided in \ref{app:distrTumorClassesTrain}.

To examine our network's behaviour, we first evaluate it on the training data of the challenge. For this, we run a 5-fold cross validation where each fold contains both HG and LG images. We then retrain the network using all training images, before applying it on the test data.

\textbf{CRF configuration:} For the multi-class problem it is challenging to find a global set of parameters for the CRF which can consistently improve the segmentation of all classes. So instead we merge the four predicted probability maps into a single \quot{whole tumor} map for CRF post-processing. The CRF then only refines the boundaries between tumor and background and additionally removes isolated false positives. Similarly to the experiments on TBI, the CRF is configured on a random subset of 44 HG and 18 LG training images, which are then reshuffled into the subsequent 5-fold cross validation. 

\subsubsection{Results}
\label{subsubsec:resBrats2015}

\begin{table}[!h]
\centering
\scriptsize
\caption{Average performance of our system on the training data of BRATS 2015 as computed on the online evaluation platform and comparison to other submissions visible at the time of manuscript submission. Presenting only teams that submitted more than half of the 274 cases. Numbers in bold indicate significant improvement by the CRF, according to a two-sided, paired t-test on the DSC metric (*$p<5\cdot 10^{-2}$, **$p<10^{-3}$). }
\label{table:onlineEvalBrats2015Training}
\begin{tabular}{@{}lllllllllll@{}}
\toprule
              & \multicolumn{3}{c}{DSC}  & \multicolumn{3}{c}{Precision} & \multicolumn{3}{c}{Sensitivity} &       \\ \cmidrule(lr){2-10}
              	& Whole & Core 	& Enh. 		& Whole   & Core  	& Enh.  & Whole & Core & Enh.   	& Cases \\ \midrule
              
Ensemble+CRF		& \textbf{90.1}*	&75.4	& \textbf{72.8}*	& 91.9	& 85.7	& 75.5	& 89.1	& 71.7	& 74.4	&274 \\
Ensemble			& 90.0			&75.5	& 72.8			& 90.3	& 85.5	& 75.4	& 90.4	& 71.9	& 74.3	&274 \\
DeepMedic+CRF	& \textbf{89.8}**&75.0	& \textbf{72.1}*	& 91.5	& 84.4	& 75.9	& 89.1	& 72.1	& 72	.5	&274 \\
DeepMedic		& 89.7			& 75.0	& 72.0			& 89.7	& 84.2	& 75.6	& 90.5	& 72.3	& 72.5	&274 \\

bakas1		 	& 88				& 77		& 68				& 90		& 84		& 68		& 89		& 76		& 75		&186\\
peres1		 	& 87				& 73		& 68				& 89		& 74		& 72		& 86		& 77		& 70		&274\\
anon1		 	& 84				& 67		& 55				& 90		& 76		& 59		& 82		& 68		& 61		&274\\
thirs1		 	& 80				& 66		& 58				& 84		& 71		& 53 	& 79		& 66		& 74		&267\\
peyrj			& 80				& 60		& 57				& 87		& 79		& 59		& 77		& 53		& 60		&274\\
\bottomrule
\end{tabular}
\end{table}

\begin{figure}[!h]
\centering
\begin{subfigure}[b]{1.0\textwidth}
	\centering
	\includegraphics[clip=true, trim=0pt 0pt 0pt 0pt, width=1.0\textwidth]{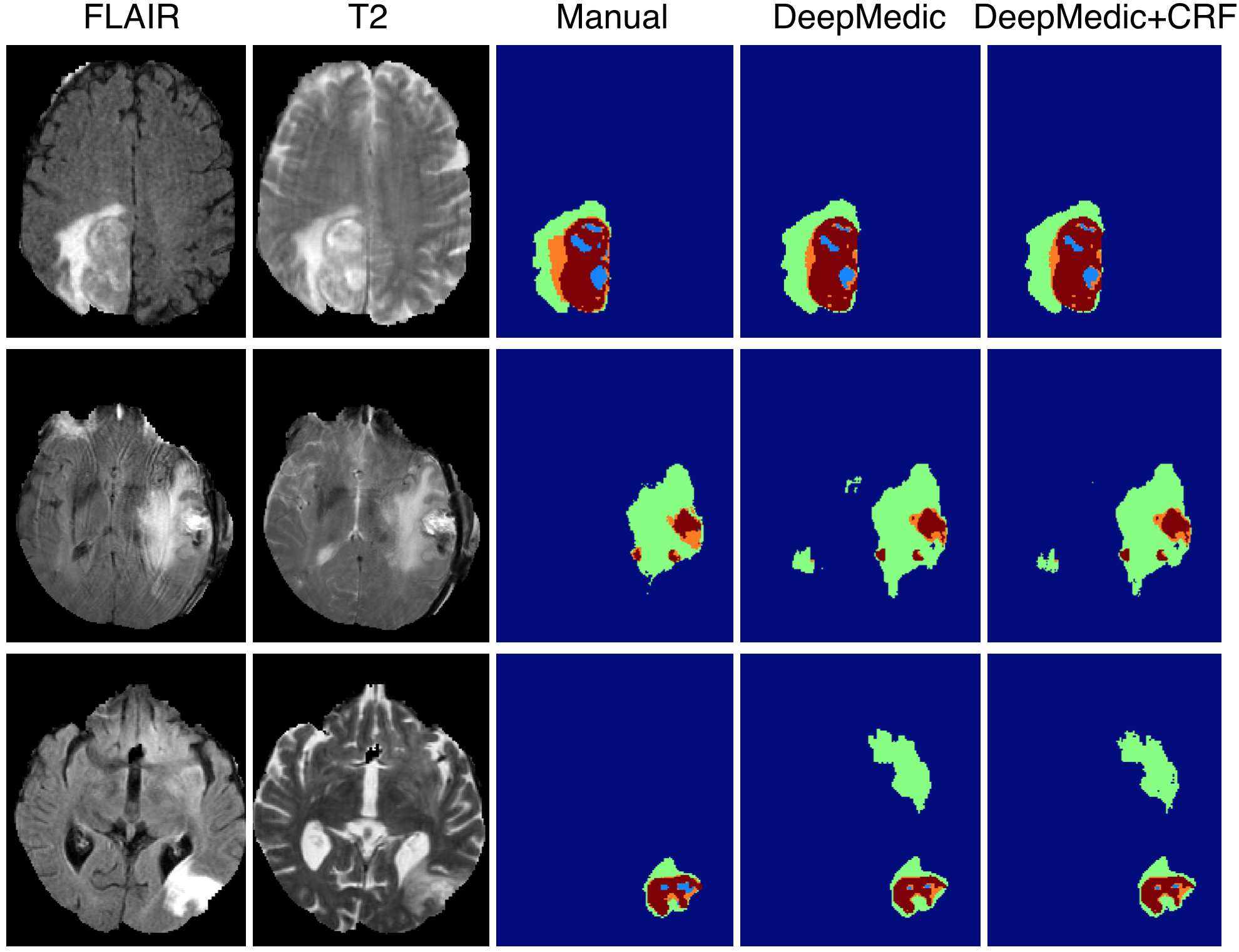}
\end{subfigure}
\vspace{-0pt} 
\caption{Examples of DeepMedic's segmentation from its evaluation on the training datasets of BRATS 2015. cyan: necrotic core, green: oedema, orange: non-enhancing core, red: enhancing core. (top and middle) Satisfying segmentation of the tumor, regardless motion artefacts in certain sequences. (bottom) One of the worst cases of over-segmentation observed. False segmentation of FLAIR hyper-intensities as oedema constitutes the most common error of DeepMedic.}
\label{fig:evalBratsVisualQuality}
\end{figure}

Quantitative results from the application of the DeepMedic, the CRF and an ensemble of three similar networks on the training data are presented in Table \ref{table:onlineEvalBrats2015Training}. The latter two offer an improvement, albeit fairly small since the performance of DeepMedic is already rather high in this task. Also shown are results from previous works, as reported on the online evaluation platform. Various settings may vary among submissions, such as the pre-processing pipeline or the number of folds used for cross-validation. Still it appears that our system performs favourably compared to previous state-of-the-art, including the semi-automatic system of \cite{bakas2015Brats} (bakas1) who won the latest challenge and the method of \cite{pereira2015Brats} (peres1), which is based on grade-specific 2D CNNs and requires visual inspection of the tumor and identification of the grade by the user prior to segmentation. Examples of segmentations obtained with our method are shown in Fig.~\ref{fig:evalBratsVisualQuality}. DeepMedic behaves very well in preserving the hierarchical structure of the tumor, which we account to the large context processed by our multi-scale network.

Table~\ref{table:onlineEvalBrats2015Testing} shows the results of our method on the BRATS test data. Results of other submissions are not accessible. The decrease in performance is possibly due to the the inclusion of test images that vary significantly from the training data, such as cases acquired in clinical centers that did not provide any of the training images, something that was confirmed by the organisers. Note that performance gains obtained with the CRF are larger in this case. This indicates not only that its configuration has not overfitted to the training database but also that the CRF is robust to factors of variation between acquisition sites, which complements nicely the more sensitive CNN.

\begin{table}[!h]
\centering
\scriptsize
\caption{Average performance of our system on the 110 test cases of BRATS 2015, as computed on the online evaluation platform. Numbers in bold indicate significant improvement by the CRF, according to a two-sided, paired t-test on the DSC metric (*$p<5\cdot 10^{-2}$, **$p<10^{-3}$). The decrease of the mean DSC by the CRF and the ensemble for the \quot{Core} class was not found significant.}
\label{table:onlineEvalBrats2015Testing}
\begin{tabular}{@{}llllllllll@{}}
\toprule
              & \multicolumn{3}{c}{DSC}  & \multicolumn{3}{c}{Precision} & \multicolumn{3}{c}{Sensitivity} \\ \cmidrule(l){2-10} 
              & Whole 			& Core & Enh. 			& Whole   & Core   & Enh.	& Whole    & Core   & Enh.   \\ \midrule

DeepMedic     & 83.6  			& 67.4 & 62.9      		& 82.3    & 84.6   & 64.0    & 88.5     & 61.6   & 65.6      \\
DeepMedic+CRF & \textbf{84.7}** 	& 67.0 & 62.9      		& 85.0    & 84.8   & 63.4    & 87.6     & 60.7   & 66.2      \\
Ensemble      & 84.5  			& 66.7 & 63.3      		& 83.3    & 86.1   & 63.2    & 88.9     & 59.9   & 67.3      \\
Ensemble+CRF  & \textbf{84.9}** 	& 66.7 & \textbf{63.4}* 	& 85.3    & 86.1   & 63.4    & 87.7     & 60.0   & 67.4		\\
\bottomrule
\end{tabular}
\end{table}

\subsection{Ischemic Stroke Lesion Segmentation}
\label{subsec:evalIsles}

\subsubsection{Material and Pre-Processing}

We participated in the 2015 Ischemic Stroke Lesion Segmentation (ISLES) challenge, where our system achieved the best results among all participants on sub-acute ischemic stroke lesions (\cite{maier2017isles}). In the training phase of the challenge, 28 datasets have been made available, along with manual segmentations. Each dataset included T1, T1-contrast, FLAIR and DWI sequences. All images were provided as skull-stripped and resampled to isotropic $1mm^3$ voxel resolution. Each volume is of size 230$\times$230$\times$154. In the testing stage, teams were provided with 36 datasets for evaluation. The test data were acquired in two clinical centers, with one of them being the same that provided all training images. Corresponding expert segmentations were hidden and results had to be submitted to an online evaluation platform. Similar to BRATS, the only pre-processing that we applied is the normalization of each image to the zero-mean and unit variance.

\subsubsection{Experimental Setting}

\textbf{Network Configuration and Training:} The configuration of the network employed is described in \cite{kamnitsas2015Isles}. The main difference with the configuration used for TBI and tumors as employed above is the relatively smaller number of FMs in the low-resolution pathway. This choice should not significantly influence accuracy on the generally small SISS lesions but it allowed us to lower the computational cost.

Similar to the other experiments, we evaluate our network with a 5-fold cross validation on the training datasets. We use data augmentation with sagittal reflections. For the testing phase of the challenge, we trained an ensemble of three networks on all training cases and aggregate their predictions by averaging.

\textbf{CRF configuration:} The parameters of the CRF were configured via a random search on the whole training dataset.

\subsubsection{Results}
\label{subsubsec:resIsles2015}

\begin{figure}[!h]
\centering
\begin{subfigure}[b]{1.0\textwidth}
	\centering
	\includegraphics[clip=true, trim=0pt 0pt 0pt 0pt, width=1.0\textwidth]{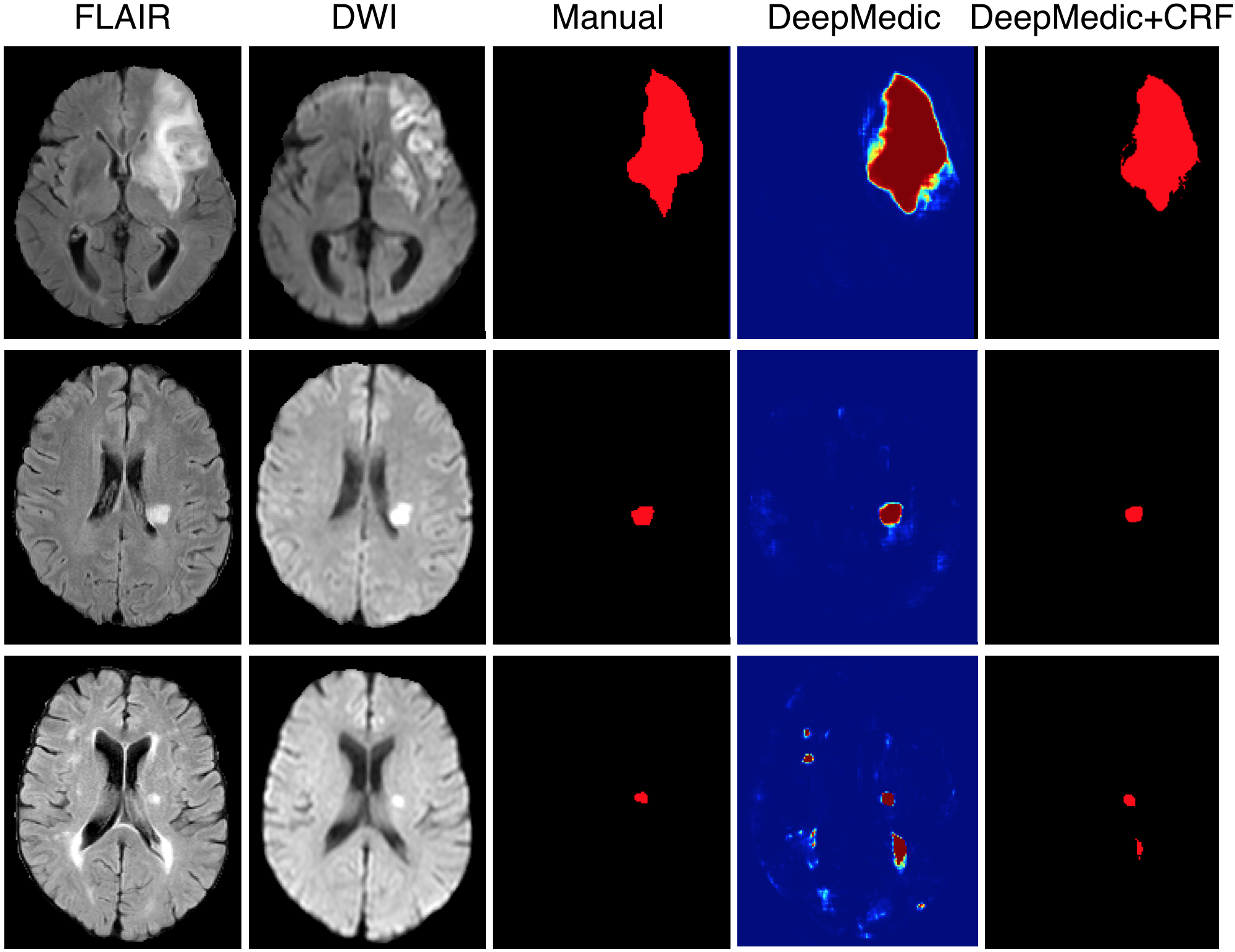}
\end{subfigure}
\vspace{-0pt} 
\caption{Examples of segmentations performed by our system on the training datasets of (SISS) ISLES 2015. (top and middle) The system is capable of satisfying segmentation of both large and smaller lesions. (bottom) Common mistakes are performed due to the challenge of differentiating stroke lesions from White Matter lesions. }
\label{fig:evalIslesVisualQuality}
\end{figure}

The performance of our system on the training data is shown in Table~\ref{table:accuracyIslesTraining}. Significant improvement is achieved by the structural regularisation offered by the CRF, although it could be partially accounted for by overfitting the training data during the CRF's configuration. Examples for visual inspection are shown in Fig.~\ref{fig:evalIslesVisualQuality}.

\begin{table}[!h]
\centering
\scriptsize
\caption{Performance of our system on the training data of the ISLES-SISS 2015 competition. Values correspond to the mean (and standard deviation). Numbers in bold indicate significant improvement by the CRF, according to a two-sided, paired t-test on the DSC metric ($p<10^{-2}$).}
\label{table:accuracyIslesTraining}
\begin{tabular}{@{}llllll@{}}
\toprule
\multicolumn{1}{c}{}		& DSC				& Precision		& Sensitivity	& ASSD			& Haussdorf 	\\ \midrule
DeepMedic				& 64(23)		 		& 68(24)			& 65(23)			& 6.99(9.91)		& 73.32(26.03)	\\
DeepMedic+CRF			& \textbf{66(24)}	& 77(24)			& 63(25)			& 5.00(10.33	)	& 55.93(28.55)	\\
\bottomrule
\end{tabular}
\end{table}

\begin{table}[!h]
\centering
\scriptsize
\caption{Our ensemble of three networks, coupled with the fully connected CRF obtained overall best performance among all participants in the testing stage of the ISLES-SISS 2015 challenge. Shown is the performance of our pipeline along with the second and third entries. Values correspond to the mean (and standard deviation).}
\label{table:accuracyIslesTesting}
\begin{tabular}{@{}llllll@{}}
\toprule
\multicolumn{1}{c}{}		& DSC		& Precision		& Sensitivity	& ASSD			& Haussdorf 	\\ \midrule
kamnk1(ours)				& 59(31)		& 68(33)			& 60(27) 		& 7.87(12.63)	& 39.61(30.68)	\\
fengc1					& 55(30)		& 64(31)			& 57(33)	 		& 8.13(15.15)	& 25.02(22.02)	\\
halmh1					& 47(32)		& 47(34)			& 56(33)	 		& 14.61(20.17)	& 46.26(34.81)	\\
\bottomrule
\end{tabular}
\end{table}

For the testing phase of the challenge we formed an ensemble of three networks, coupled with the fully connected CRF. Our submission ranked first, indicating superior performance on this challenging task among 14 submissions. Table~\ref{table:accuracyIslesTesting} shows our results, along with the other two top entries (\cite{feng2015Isles,halme2015Isles}). Among the other participating methods was the CNN of \cite{Havei2015Journal} with 3 layers of 2D convolutions. That method perfomed less well on this challenging task (\cite{maier2017isles}). This points out the advantage offered by 3D context, the large field of view of DeepMedic thanks to multi-scale processing and the representational power of deeper networks. It is important to note the decrease of performance in comparison to the training set. All methods performed worse on the data coming from the second clinical center, including the method of \cite{feng2015Isles} that is not machine-learning based. This highlights a general difficulty with current approaches when applied on multi-center data.

\subsection{Implementation Details}
 
Our CNN is implemented using the Theano library (\cite{Bastien-Theano-2012}). Each training session requires approximately one day on an NVIDIA GTX Titan X GPU using cuDNN v5.0. The efficient architecture of DeepMedic also allows models to be trained on GPUs with only 3GB of memory. Note that although dimensions of the volumes in the processed databases do not allow dense training on whole volumes for this size of network, dense inference on a whole volume is still possible, as it requires only a forward-pass and thus less memory. In this fashion segmentation of a volume takes less than 30 seconds but requires 12 GB of GPU memory. Tiling the volume into multiple segments of size $35^3$ allows inference on 3 GB GPUs in less than three minutes.

Our 3D fully connected CRF is implemented by extending the original source code by \cite{Krahenbuhl2013}. A CPU implementation is fast, capable of processing a five-channel brain scan in under three minutes. Further speed-up could be achieved with a GPU implementation, but was not found necessary in the scope of this work.



\section{Discussion and Conclusion}
\label{sec:discussion}

We have presented DeepMedic, a 3D CNN architecture for automatic lesion segmentation that surpasses state-of-the-art on challenging data. The proposed novel training scheme is not only computationally efficient but also offers an adaptive way of partially alleviating the inherent class-imbalance of segmentation problems. We analyzed the benefits of using small convolutional kernels in 3D CNNs, which allowed us to develop a deeper and thus more discriminative network, without increasing the computational cost and number of trainable parameters. We discussed the challenges of training deep neural networks and the adopted solutions from the latest advances in deep learning. Furthermore, we proposed an efficient solution for processing large image context by the use of parallel convolutional pathways for multi-scale processing, alleviating one of the main computational limitations of previous 3D CNNs. Finally, we presented the first application of a 3D fully connected CRF on medical data, employed as a post-processing step to refine the network's output, a method that has also been shown promising for processing 2D natural images (\cite{chen2014semantic}). The design of the proposed system is well suited for processing medical volumes thanks to its generic 3D nature. The capabilities of DeepMedic and the employed CRF for capturing 3D patterns exceed those of 2D networks and locally connected random fields, models that have been commonly used in previous work. At the same time, our system is very efficient at inference time, which allows its adoption in a variety of research and clinical settings.

The generic nature of our system allows its straightforward application for different lesion segmentation tasks without major adaptations. To the best of our knowledge, our system achieved the highest reported accuracy on a cohort of patients with severe TBI. As a comparison, we improved over the reported performance of the pipeline in \cite{Rao2014b}. Important to note is that the latter work focused only on segmentation of contusions, while our system has been shown capable of segmenting even small and diffused pathologies. Additionally, our pipeline achieved state-of-the-art performance on both public benchmarks of brain tumors (BRATS 2015) and stroke lesions (SISS ISLES 2015). We believe performance can be further improved with task- and data-specific adjustments, for instance in the pre-processing, but our results show the potential of this generically designed segmentation system.

When applying our pipeline to new tasks, a laborious process is the reconfiguration of the CRF. The model improved our system's performance with statistical significance in all investigated tasks, most profoundly when the performance of the underlying classifier degrades, proving its flexibility and robustness. Finding optimal parameters for each task, however, can be challenging. This became most obvious on the task of multi-class tumor segmentation. Because the tumor's substructures vary significantly in appearance, finding a global set of parameters that yields improvements on all classes proved difficult. Instead, we applied the CRF in a binary fashion. This CRF model can be configured with a separate set of parameters for each class. However the larger parameter space would complicate its configuration further. Recent work from \cite{Zheng2015} showed that this particular CRF can be casted as a neural network and its parameters can be learned with regular gradient descent. Training it in an end-to-end fashion on top of a neural network would alleviate the discussed problems. This will be explored as part of future work.

\begin{figure}[!h]
\vspace{-20pt}
\centering
\begin{subfigure}[b]{0.95\textwidth}
	\centering
	\includegraphics[clip=true, trim=0pt 0pt 0pt 0pt, width=1.0\textwidth]{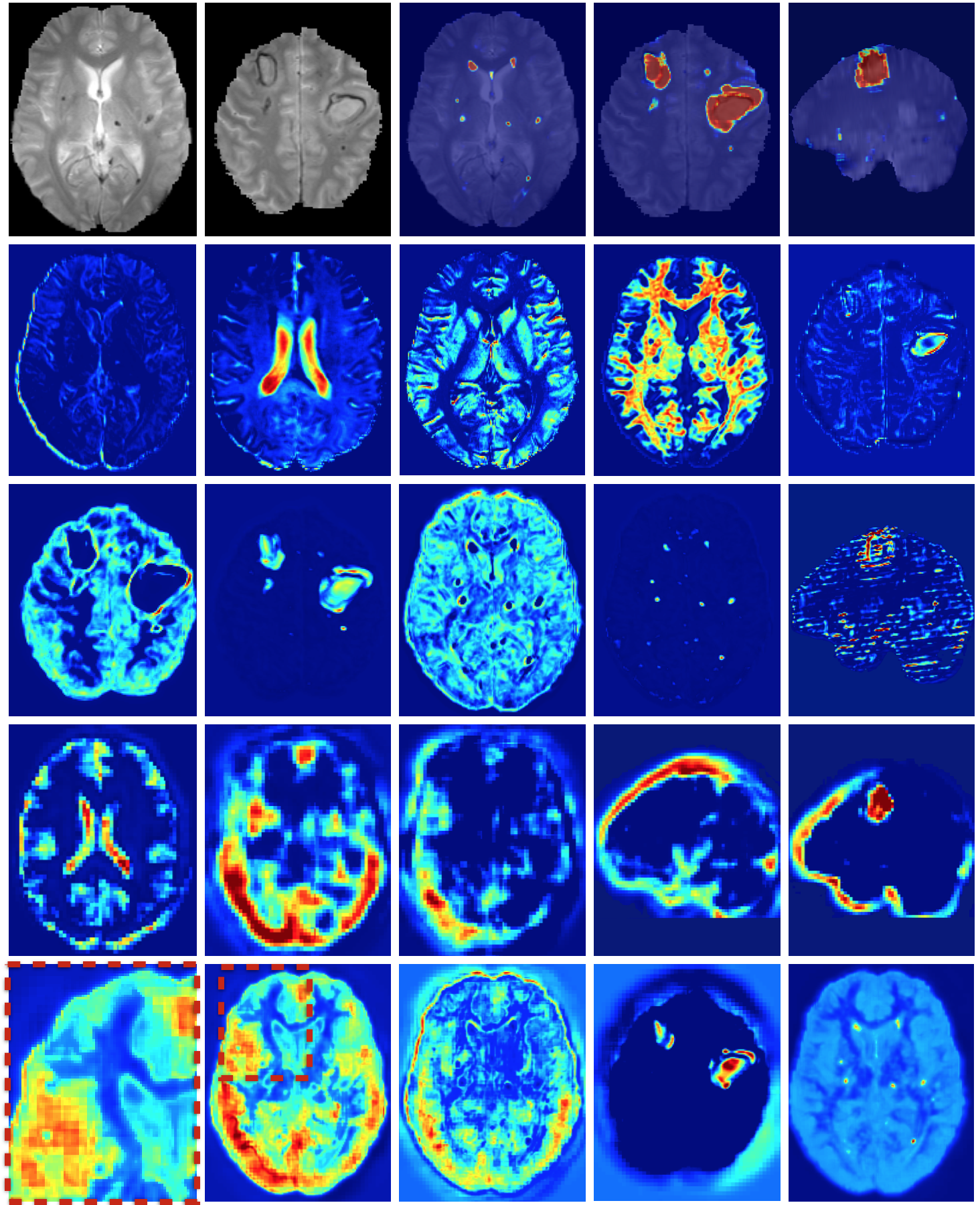}
\end{subfigure}
\vspace{-5pt} 
\caption{(First row) GE scan and DeepMedic's segmentation. (Second row) FMs of earlier and (third row) deeper layers of the first convolutional pathway. (Fourth row) Features learnt in the low-resolution pathway. (Last row) FMs of the two last hidden layers, which combine multi-resolution features towards the final segmentation.}
\label{fig:featureMaps}
\end{figure}

The discriminative power of the learned features is indicated by the success of recent CNN-based systems in matching human performance in domains where it was previously considered too ambitious (\cite{he2015delving, Silver2016}). Analysis of the automatically extracted information could potentially provide novel insights and facilitate research on pathologies for which little prior knowledge is currently available. In an attempt to illustrate this, we explore what patterns have been learned automatically for the lesion segmentation tasks. We visualize the activations of DeepMedic's FMs when processing a subject from our TBI database. Many appearing patterns are difficult to interpret, especially in deeper layers. In Fig.~\ref{fig:featureMaps} we provide some examples that have an intuitive explanation. One of the most interesting findings is that the network learns to identify the ventricles, CSF, white and gray matter. This reveals that differentiation of tissue type is beneficial for lesion segmentation. This is in line with findings in the literature, where segmentation performance of traditional classifiers was significantly improved by incorporation of tissue priors (\cite{Leemput1999, Zikic2012}). It is intuitive that different types of lesions affect different parts of the brain depending on the underlying mechanisms of the pathology. A rigorous analysis of spatial cues extracted by the network may reveal correlations that are not well defined yet.

Similarly intriguing is the information extracted in the low-resolution pathway. As they process greater context, these neurons gain additional localization capabilities. The activations of certain FMs form fields in the surrounding areas of the brain. These patterns are preserved in the deepest hidden layers, which indicates they are beneficial for the final segmentation (see two last rows of Fig.~\ref{fig:featureMaps}). We believe these cues provide a spatial bias to the system, for instance that large TBI contusions tend to occur towards the front and sides of the brain (see Fig.~\ref{fig:spatialMap}). Furthermore, the interaction of the multi-resolution features can be observed in FMs of the hidden layer that follows the concatenation of the pathways. The network learns to weight the output of the two pathways, preserving low resolution in certain parts and show fine details in others (bottom row of Fig.~\ref{fig:featureMaps}, first three FMs). Our assumption is that the low-resolution pathway provides a rough localization of large pathologies and brain areas that are challenging to segment, which reserves the rest of the network's capacity for learning detailed patterns associated with the detection of smaller lesions, fine structures and ambiguous areas.

The findings of the above exploration lead us to believe that great potential lies into fusing the discriminative power of the \quot{deep black box} with the knowledge acquired over years of targeted biomedical research. Clinical knowledge is available for certain pathologies, such as spatial priors for white matter lesions. Previously engineered models have been proven effective in tackling fundamental imaging problems, such as brain extraction, tissue segmentation and bias field correction. We show that a network is capable of automatically extracting some of this information. It would be interesting, however, to investigate structured ways for incorporating such existing information as priors into the network's feature space, which should simplify the optimization problem while letting a specialist guide the network towards an optimal solution.

Although neural networks seem promising for medical image analysis, making the inference process more interpretable is required. This would allow understanding when the network fails, an important aspect in biomedical applications. Although the output is bounded in the $[0,1]$ range and commonly referred to as probability for convenience, it is not a true probability in a Bayesian sense. Research towards Bayesian networks aims to alleviate this limitation. An example is the recent work of \cite{Gal2015} who show that model confidence can be estimated via sampling the dropout mask.

A general point should be made about the performance drop observed when our system is applied on test datasets of BRATS and ISLES in comparison to its cross-validated performance on the training data. In both cases, subsets of the test images were acquired in clinical centers different from the ones of training datasets. Differences in scanner type and acquisition protocols have significant impact on the appearance of the images. The issue of multi-center data heterogeneity is considered a major bottleneck for enabling large-scale imaging studies. This is not specific to our approach, but a general problem in medical image analysis. One possible way of making the CNN invariant to the data heterogeneity is to learn a generative model for the data acquisition process, and use this model in the data augmentation step. This is a direction we explore as part of future work.

In order to facilitate further research in this area and to provide a baseline for future evaluations, we make the source code of the entire system publicly available.



\section*{Acknowledgements}

This work is supported by the EPSRC First Grant scheme (grant ref no. EP/N023668/1) and partially funded under the 7th Framework Programme by the European Commission (TBIcare: http://www.tbicare.eu/; CENTER-TBI: https://www.center-tbi.eu/).
This work was further supported by a Medical Research Council (UK) Program Grant (Acute brain injury: heterogeneity of mechanisms, therapeutic targets and outcome effects [G9439390 ID 65883]), the UK National Institute of Health Research Biomedical Research Centre at Cambridge and Technology Platform funding provided by the UK Department of Health. KK is supported by the Imperial College London PhD Scholarship Programme. VFJN is supported by a Health Foundation/Academy of Medical Sciences Clinician Scientist Fellowship. DKM is supported by an NIHR Senior Investigator Award. We gratefully acknowledge the support of NVIDIA Corporation with the donation of two Titan X GPUs for our research.


\appendix


\section{Additional Details on Multi-Scale Processing}
\label{app:detailsMultiscale}

The integration of multi-scale parallel pathways in architectures that use solely unary kernel strides, such as the proposed, was described in Sec.~\ref{subsec:multiscaleCnn}. The required up-sampling of the low-resolution features was performed with simple repetition in our experiments. This was found sufficient, with the following hidden layers learning to combine the multi-scale features. In the case of architectures with strides greater than unary, the last convolutional layers of the two pathways, $L1$ and $L2$, have receptive fields $\boldsymbol{\varphi}_{L1}$ and $\boldsymbol{\varphi}_{L2}$ with strides $\boldsymbol{\tau}_{L1}$ and $\boldsymbol{\tau}_{L2}$ respectively. To preserve spatial correspondence of the multi-scale features and enable the network for dense inference, the dimensions of the input segments should be chosen such that the FMs in $L2$ can be brought to the dimensions of the FMs in $L1$ after sequential resampling by $\uparrow \boldsymbol{\tau}_{L2}$, $\uparrow F_D$, $\downarrow \boldsymbol{\tau}_{L1}$ or equivalent combinations. Here $\uparrow$ and $\downarrow$ represent up- and down-sampling by the given factor. Because they are more reliant on these operations, utilization of more elaborate, learnt upsampling schemes (\cite{Long2014, Ronneberger2015, Noh2015}) should be beneficial in such networks.

\section{Additional Details on Network Configurations}
\label{app:detailsConfig}

\textbf{3D Networks:} The main description of our system is presented in Sec.~\ref{sec:segmentationSystem}. All models discussed in this work outside Sec.~\ref{subsec:val3dContext} are fully 3D CNNs. Their architectures are presented in Table \ref{subtab:netsConfig3d}. They all use the PReLu non-linearity (\cite{he2015delving}). They are trained using the RMSProp optimizer (\cite{rmsProp}) and Nesterov momentum (\cite{sutskever2013importance}) with value $m=0.6$. $L1 = 10^{-6}$ and $L2 = 10^{-4}$ regularisation is applied. We train the networks with dense-training on batches of 10 segments, each of size $25^3$. Exceptions are the experiments in Sec~\ref{subsec:valDenseTraining}, where the batch sizes were adjusted along with the segment sizes, to achieve similar memory footprint and training time per batch. The weights of our shallow, 5-layers networks are initialized by sampling from a normal distribution $\mathcal{N}(0,0.01)$ and their initial learning rate is set to $a=10^{-4}$. Deeper models (and the \quot{Shallow+} model in Sec~\ref{subsec:valDeeper}) use the weight initialisation scheme of \cite{he2015delving}. The scheme increases the signal's variance in our settings, which leads to RMSProm decreasing the effective learning rate. To counter this, we accompany it with an increased initial learning rate $a = 10^{-3}$. Throughout training, the learning rate of all models is halved whenever convergence plateaus. Dropout with 50\% rate is employed on the two last hidden layers of 11-layers deep models.

\textbf{2D Networks:} Table \ref{subtab:netsConfig2d} presents representative examples of 2D configurations that were employed for the experiments discussed in Sec.~\ref{subsec:val3dContext}. Width, depth and batch size were adjusted so that total required memory was similar to the 3D version of DeepMedic. Wider or deeper variants than the ones presented did not show greater performance. A possible reason is that this number of filters is enough for the extraction of the limited 2D information and that the field of view of the deep multi-scale variant is already sufficient for the application. The presented 2D models were regularized with $L1 = 10^{-8}$ and $L2 = 10^{-6}$ since they have less parameters than the 3D variants. All but Dm2dPatch were trained with momentum $m=0.6$ and initial learning rate $a = 10^{-3}$, while the rest with $m=0.9$ and $a = 10^{-2}$ as this setting increased performance. The rest of the hyper parameters are the same as for the 3D DeepMedic.

\setcounter{table}{0}    
\renewcommand\thetable{B.\arabic{table}}

\begin{table}[!h]
\centering
\scriptsize
\caption{Network architectures investigated in Sec.~\ref{sec:vaOfNetArch} and final validation accuracy achieved in the corresponding experiments. (a) 3D and (b) 2D architectures. Columns from left to right: model's name, number of parallel identical pathways and number of feature maps at each of their convolutional layers, number of feature maps at each hidden layer that follows the concatenation of the pathways, dimensions of input segment to the normal and low resolution pathways, batch size and, finally, average DSC achieved on the validation fold. Further configuration details provided in \ref{app:detailsConfig}.}
\label{tab:netsConfig}
\begin{subtable}{1.0\linewidth}
\caption{3D Network Architectures}
\label{subtab:netsConfig3d}
\begin{tabular}{@{}m{1.5cm}m{3.7cm}m{1.2cm}m{1.2cm}m{1.2cm}m{0.8cm}m{1.3cm}}
\toprule	
	               & \#Pathways: FMs/Layer       & FMs/Hidd. & Seg.Norm. & Seg.Low &B.S. & DSC(\%)    \\ \midrule
Shallow(+)         & 1: 30,40,40,50                  & -          & 25x25x25   & -        &10  & 60.2(61.7) \\
Deep(+)            & 1: 30,30,40,40,40,40,50,50      & -          & 25x25x25   & -        &10  & 00.0(64.9)  \\
BigDeep+           & 1: 60,60,80,80,80,80,100,100    & 150,150    & 25x25x25   & -        &10  & 65.2       \\
DeepMedic          & 2: 30,30,40,40,40,40,50,50      & 150,150    & 25x25x25   & 19x19x19 &10  & 66.6       \\ \bottomrule
\end{tabular}
\end{subtable}%
\vspace{10pt}
\begin{subtable}{1.0\linewidth}
\caption{2D Network Architectures}
\label{subtab:netsConfig2d}
\begin{threeparttable}
\begin{tabular}{@{}m{1.5cm}m{3.7cm}m{1.2cm}m{1.2cm}m{1.2cm}m{0.8cm}m{1.3cm}}
\toprule	
	            & \#Pathways: FMs/Layer       & FMs/Hidd. & Seg.Norm. & Seg.Low &B.S. & DSC(\%)    \\ \midrule
Dm2dPatch*    	& 2: 30,30,40,40,40,40,50,50      & 150,150    & 17x17x1    & 17x17x1    &540 & 58.8       \\
Dm2dSeg        & 2: 30,30,40,40,40,40,50,50      & 150,150    & 25x25x1    & 19x19x1    &250 & 60.9       \\
Wider2dSeg     & 2: 60,60,80,80,80,80,100,100    & 200,200    & 25x25x1    & 19x19x1    &100 & 61.3       \\
Deeper2dSeg    & 2: 16 layers, linearly 30 to 50 & 150,150    & 41x41x1    & 35x35x1    &100 & 61.5       \\
Large2dSeg  	& 2: 12 layers, linearly 45 to 80 & 200,200    & 33x33x1    & 27x27x1    &100 & 61.3    \\ \bottomrule
\end{tabular}
\begin{tablenotes}
            \item[*] Sampling was manually calibrated to achieve similar class balance as models that are trained on image segments. Model underperformed otherwise.
\end{tablenotes}
\end{threeparttable}
\end{subtable}
\end{table}

\section{Distribution of Tumor Classes Captured in Training}
\label{app:distrTumorClassesTrain}
\setcounter{table}{0}    
\renewcommand\thetable{C.\arabic{table}} 

\hyperref[table:trainingSamplesPercBrats2015Training]{Table C.1}

\begin{table}[!h]
\centering
\scriptsize
\caption{Real distribution of the classes in the training data of BRATS 2015, along with the distribution captured by our proposed training scheme, when segments of size $25^3$ are extracted centred on the tumor and healthy tissue with equal probability. Relative distribution of the foreground classes is closely preserved and the imbalance in comparison to the healthy tissue is automatically alleviated.}
\label{table:trainingSamplesPercBrats2015Training}
\begin{tabular}{@{}lccccc@{}}
\toprule
\multicolumn{1}{c}{} & Healthy		& Necrosis 	& Edema 		& Non-Enh. 	& Enh.Core 	\\ \midrule
Real		 			& 92.42			& 0.43		& 4.87		& 1.02		& 1.27		\\
Captured				& 58.65			& 2.48		& 24.98		& 6.40		& 7.48		\\
\bottomrule
\end{tabular}
\end{table}

\bibliographystyle{elsarticle-harv}
\bibliography{journalDeepMedic}

\end{document}